\RequirePackage{fix-cm}
\documentclass[twocolumn]{svjour3}          
\smartqed  
\usepackage{graphicx}
\usepackage{subfigure}
\usepackage{times}
\usepackage{booktabs} 
\usepackage{amsmath}
\usepackage{color}
\usepackage{cite}
\usepackage{algorithm}
\usepackage{algorithmic}
\usepackage{natbib}
\usepackage[dvipsnames]{xcolor}
\usepackage{amssymb}
\usepackage{array}
\usepackage{bbm}
\usepackage{enumitem}
\usepackage{calc}
\usepackage{multirow}
\usepackage{xspace}
\usepackage{caption}
\usepackage[utf8]{inputenc}
\usepackage{epsfig}
\usepackage{tabularx}
\usepackage{threeparttable}
\usepackage{colortbl}
\usepackage{hhline}
\usepackage[misc]{ifsym}

\newcommand{\ie}{\textit{i.e.}}
\newcommand{\eg}{\textit{e.g}}

\usepackage[dvipsnames]{xcolor}
\newcommand{\stdvu}[1]{\scriptsize{\color{darkgray}(#1)} {\color{ForestGreen}$\uparrow$}}

\definecolor{mypink}{rgb}{.99,.91,.95}
\usepackage{etoolbox}
\makeatletter
\usepackage[dvipsnames]{xcolor}

\newcommand{\stdvd}[1]{\scriptsize{\color{darkgray}(#1)} {\color{red}$\downarrow$}}

\newcommand{\stdvno}[1]{\scriptsize{\color{darkgray}(#1)} {\color{mygray}$\downarrow$}}
\definecolor{firebrick}{rgb}{0.7, 0.13, 0.13}
\definecolor{darkpastelgreen}{rgb}{0.01, 0.75, 0.24}
\definecolor{deepskyblue}{rgb}{0.0, 0.75, 1.0}
\definecolor{mypink2}{rgb}{.99,.96,.98}
\definecolor{mypink1}{rgb}{.99,.93,.98}
\definecolor{mypink}{rgb}{.99,.90,.98}
\definecolor{mygray}{rgb}{.95,.95,.95}
\definecolor{lv14}{rgb}{0.5,0.5,0.5}
\definecolor{DarkSeaGreen1}{rgb}{0.56, 0.93, 0.56}
\definecolor{LightBlue1}{rgb}{0.75, 1, 1}
\let\@algcomment\relax

\newcommand\algcomment[1]{\def\@algcomment{\footnotesize#1}}
\renewcommand\fs@ruled{\def\@fs@cfont{\bfseries}\let\@fs@capt\floatc@ruled
	\def\@fs@pre{\hrule height.8pt depth0pt \kern2pt}%
	\def\@fs@post{}%
	\def\@fs@mid{\kern2pt\hrule\kern2pt}%
	\let\@fs@iftopcapt\iftrue}
\makeatother
\usepackage{tabulary}
\newcolumntype{I}{!{\vrule width 1pt}}

\newcolumntype{x}[1]{>{\centering\arraybackslash}p{#1pt}}
\newcolumntype{y}[1]{>{\raggedright\arraybackslash}p{#1pt}}
\newcolumntype{z}[1]{>{\raggedleft\arraybackslash}p{#1pt}}
\newlength\savewidth

\journalname{International Journal of Computer Vision}

\begin{document}
\begin{sloppypar}
\title{Diffusion-Enhanced Test-time Adaptation with Text and Image Augmentation
}

\author{Chun-Mei Feng  \and
Yuanyang He \and Jian Zou \and Salman Khan \and Huan Xiong \\\and Zhen Li $^{(\textrm{\Letter})}$ \and Wangmeng Zuo \and Rick Siow Mong Goh \and Yong Liu
}
%
%
\institute{
{\textbullet{} Chun-Mei Feng, Rick Siow Mong Goh, and Yong Liu are with the Institute of High Performance Computing (IHPC), Agency for Science, Technology and Research (A*STAR), Singapore, \email{fengcm.ai@gmail.com};\\}
{\textbullet{} Yuanyang He is with the National University of Singapore, Singapore;\\}
{\textbullet{} Jian Zou, Huan Xiong and Wangmeng Zuo are with the Harbin Institute of Technology, Harbin, China;\\}
{\textbullet{} Salman Khan is with the Mohamed bin Zayed University of Artificial Intelligence (MBZUAI), UAE, and Australian National University, Canberra ACT, Australia;\\}
{\textbullet{} Zhen Li is with the Chinese University of Hong Kong, Shenzhen, China. FNii-Shenzhen.\\}
{\textbullet{} Zhen Li is the corresponding author of this work.}
}

\date{Received: date / Accepted: date}

\maketitle

\begin{abstract}
Existing test-time prompt tuning (TPT) methods focus on single-modality data, primarily enhancing images and using confidence ratings to filter out inaccurate images. 
However, while image generation models can produce visually diverse images, single-modality data enhancement techniques still fail to capture the comprehensive knowledge provided by different modalities.
Additionally, we note that the performance of TPT-based methods drops significantly when the number of augmented images is limited, which is not unusual given the computational expense of generative augmentation.
To address these issues, we introduce $\text{IT}^{3}\text{A}$, a novel test-time adaptation method that utilizes a pre-trained generative model for multi-modal augmentation of each test sample from unknown new domains.
By combining augmented data from pre-trained vision and language models, we enhance the ability of the model to adapt to unknown new test data.
Additionally, to ensure that key semantics are accurately retained when generating various visual and text enhancements, we employ cosine similarity filtering between the logits of the enhanced images and text with the original test data. 
This process allows us to filter out some spurious augmentation and inadequate combinations.
To leverage the diverse enhancements provided by the generation model across different modals, we have replaced prompt tuning with an adapter for greater flexibility in utilizing text templates.
Our experiments on the test datasets with distribution shifts and domain gaps show that in a zero-shot setting, $\text{IT}^{3}\text{A}$ outperforms state-of-the-art test-time prompt tuning methods with a $5.50$\% increase in accuracy.


\end{abstract}

\begin{figure*}[t]
	\begin{center}
	\includegraphics[width=\linewidth]{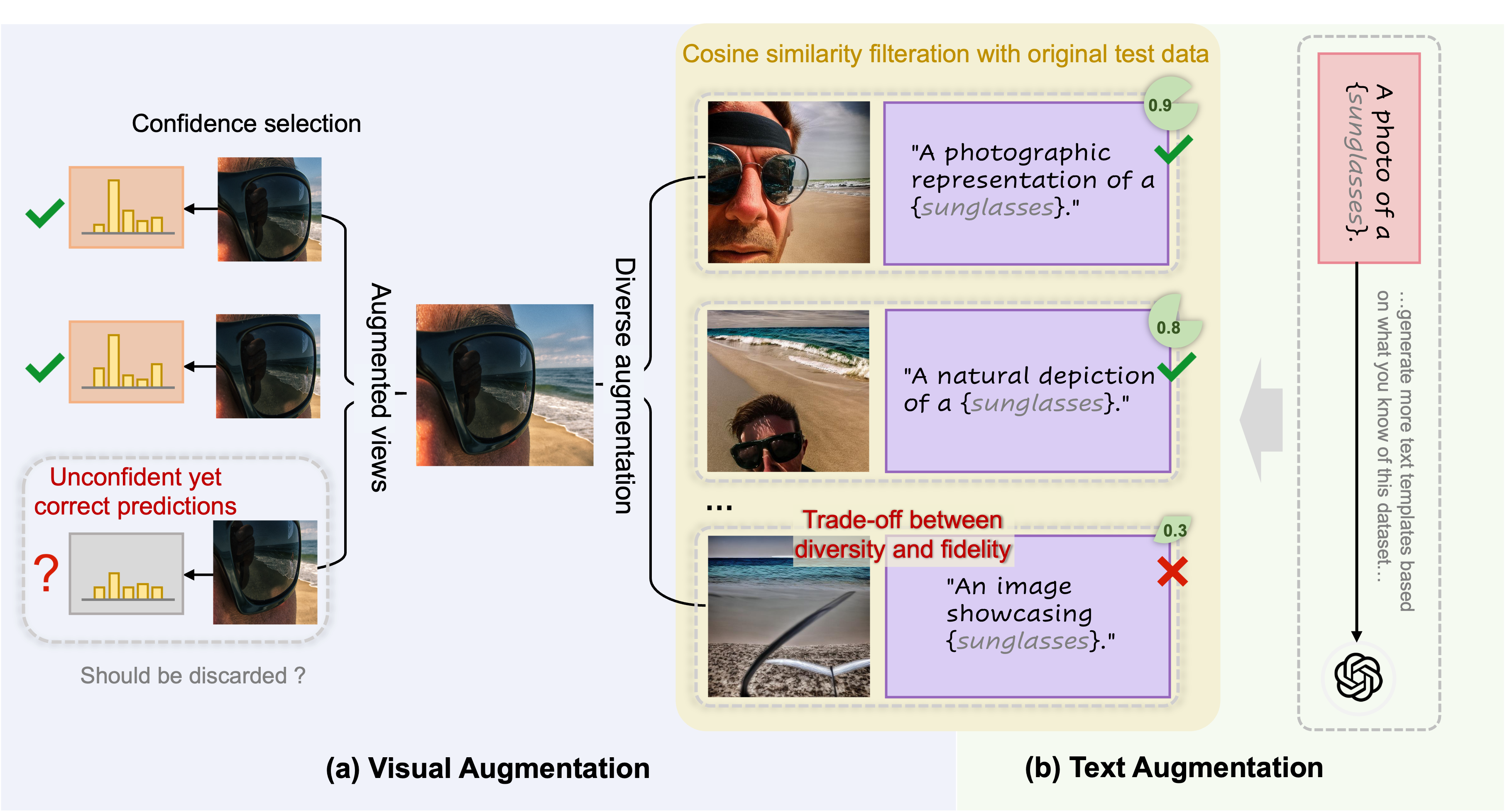}
	\end{center}
	\captionsetup{font=small}

	\caption{\small\textbf{(a)} Visual augmentation of different views (TPT~\citep{shu2022test}) and data generated through diffusion, demonstrating a richer variety of visual appearances (DiffTPT~\citep{feng2023diverse}). \textbf{(b)} Diverse text augmentation produced by GPT-4. Additionally, the augmentation from different modalities, \ie, images and text, will be combined into image-text augmentation pairs, which will then be filtered using cosine similarity to remove low-quality augmentations.}
	\label{fig:1}
\end{figure*}


\section{Introduction}\label{sec:intro}
Recent advancements have shown that pre-trained vision-language models like CLIP~\citep{radford2021learning} perform exceptionally well on a range of downstream tasks, without requiring specific task-related training data~\citep{zhou2022learning,zhou2022conditional,li2022grounded,ramesh2022hierarchical}.
Although their success is attributed to well-crafted prompts, the limitations of hand-crafted prompts and prompt tuning stem from their reliance on the training data distribution within the current domain, making it difficult to generalize to new distributions, particularly in zero-shot settings~\citep{mandal2019out}. 
To address this issue, a technique called test-time prompt tuning (TPT)~\citep{shu2022test} has been introduced, which adapts prompt embeddings for each test sample from an unseen domain in real-time, without requiring training data or annotations. This approach is more practical for dynamic real-world applications where acquiring extensive labeled data for a new target distribution is often problematic.

The early practice of TPT involves combining confidence selection with entropy minimization for prompt tuning, utilizing various augmented views of each test sample~\citep{shu2022test}. 
The augmentation method employed in~\citep{shu2022test} relies on basic parametric transformations for addressing data scarcity (see Fig.~\ref{fig:1} (a)).
However, these simple transformations 
fail to bring diversity in semantics into augmented views
~\citep{antoniou2017data,perez2017effectiveness,zhao2020differentiable,shorten2019survey}.
%
The under-diversified augmented data may result in the learned prompt fitting only to the original image details other than the key semantics, thereby compromising its generalization capability.
Moreover, the entropy-based confidence selection method proposed in~\citep{shu2022test} does not sufficiently ensure prediction accuracy, as augmented samples with low-entropy predictions can still be misclassified, producing unrepresentative samples in the augmented pool.

Recent developments in image generation technology have significantly improved the handling of varied augmented data.
Traditional image generation techniques, such as VAEs~\citep{kingma2013auto} and GANs~\citep{goodfellow2020generative}, typically necessitate large datasets for effective training.
In contrast, diffusion models have recently demonstrated exceptional capabilities in generating text-to-image outputs with impressive photo-realistic quality~\citep{nichol2021glide,ramesh2022hierarchical,saharia2022photorealistic,rombach2022high}. 
Unlike the augmentation method utilized in~\citep{shu2022test}, data produced by diffusion models can display much greater diversity, leading to richer visual representations and enhancing the generalization of the learned prompts.
However, while image generation models can produce visually different images, the information provided by unimodal data augmentation techniques remains limited.
Furthermore, continuously augmenting images to improve performance is limited by computational resources. In other words, the performance of those methods significantly drops when the number of augmented images is constrained. This is not surprising given the high computational cost of generating image augmentations.
Fortunately, with the advancement of large language models (LLMs), augmenting text using pre-trained language models to create image-text pairs with consistent content but varying styles can effectively supplement the insufficient information of the image modality.

In this work, we introduce a novel test-time adapter-based tuning method called $\text{IT}^{3}\text{A}$, which enhances data diversity for test samples using pre-trained vision and language models, thereby improving the model's adaptability to unknown new test data. We also employ cosine similarity filtering to eliminate some spurious augmentations and inadequate combinations (see Fig.~\ref{fig:1} and Fig.~\ref{fig:3.4}). 
For data diversity, $\text{IT}^{3}\text{A}$ utilizes both Stable Diffusion and GPT-4 for multi-modal data augmentation. Stable Diffusion is a text-to-image generation model that synthesizes images based on CLIP text features~\citep{rombach2022high}. Instead of using CLIP text features, we leverage the CLIP image features of the test samples and input them into Stable Diffusion for image enhancement. This diffusion-based augmentation effectively generates a variety of images with rich visual appearance changes while retaining key semantics. 
We then adopt GPT-4 with specific instructions to generate multiple text templates that vary in style but maintain consistent semantics, pairing them with the augmented images. 
Please note that the GPT-4 can be substituted by any other open-source large language model, \eg, LLaMA~\citep{touvron2023llama}, BELLE~\citep{BELLE}, Bloom~\citep{le2023bloom}, Vicuna~\cite{chiang2023vicuna}, and MOSS~\citep{Sun2024MOSS}.
To ensure prediction fidelity, we introduce cosine similarity-based filtering between the logits of the test data and the generated image-text pairs, which helps filter out spurious augmentations and inadequate combinations (see Fig.~\ref{fig:3.4}), allowing the model to strike a fair balance between diversity and fidelity. 
Experimental results indicate that, compared to state-of-the-art test-time prompt tuning methods, the $\text{IT}^{3}\text{A}$ approach improves zero-shot accuracy by an average of $5.50$\%~\citep{feng2023diverse}. In summary, our contributions are as follows:

To sum up, our contributions are as follows:
\begin{itemize}
\item 
   We propose a novel test-time adaptation method, $\text{IT}^{3}\text{A}$, that simultaneously leverages \textit{text and image augmentations}, utilizing the diverse features generated by generative models for enhanced adaptation during testing.
   %
\item 
   To ensure key semantics are faithfully preserved while generating \textit{diverse} visual and textual augmentations, we employ cosine similarity filtering between the \textit{logits} of the \textit{original} test data and the \textit{augmented images and texts}. This process removes spurious augmentation pairs, thereby improving the predictive accuracy of the enhanced images.
   %
   %
\item 
   Experimental results demonstrate that our $\text{IT}^{3}\text{A}$ method significantly outperforms state-of-the-art test-time tuning techniques.
   %
   
\end{itemize}

Our initial findings were presented at ICCV 2023~\citep{feng2023diverse}. This journal version offers three major enhancements:
\textit{Firstly}, we explore the potential of leveraging the diversity of pre-trained text generation models for test-time optimization; \eg, using GPT-4 to generate various text templates based on its pre-trained knowledge.
\textit{Secondly}, to filter out potential spurious information in the generated image and text pairs, we move the cosine similarity between the original test images and the generated images to between the logits of the original test data and the augmented images and texts, thereby removing generated data with low quality.
\textit{Thirdly}, to fully leverage the diverse augmentations of images and text provided by generative models, we have replaced prompt tuning in our conference version~\citep{feng2023diverse} with an adapter. This change allows for more flexible use of text templates on the text encoder side.
\textit{Lastly}, in addition to comparisons with state-of-the-art methods, we also conducted various ablation studies to demonstrate the effectiveness of $\text{IT}^{3}\text{A}$'s improvements.


\section{Related Work}
\subsection{Parameter-efficient Fine-tune} 
Large-scale pre-trained models have significantly boosted performance across numerous tasks in both natural language processing~\citep{devlin2018bert, radford2018improving} and computer vision~\citep{jia2021scaling, chen2020simple, jia2022visual, feng2023learning, li2022prompt}. These improvements are achieved by learning comprehensive representations and transferring the acquired knowledge to a variety of downstream applications~\cite{feng2023towards}. 
In recent years, a variety of parameter-efficient fine-tuning techniques, such as prompt tuning and adapters, have been developed to tailor pre-trained models for specific downstream tasks.
One such technique, prompt tuning, allows pre-trained models to directly adapt to downstream tasks by incorporating a small set of trainable tokens into the input space.
For example, CoOp~\citep{zhou2022learning} utilizes continuous prompt optimization, while CoCoOp~\citep{zhou2022conditional} employs instance-wise prompt conditionalization, both methods aim to improve generalization to out-of-distribution data.
Numerous studies employ adapters and non-parametric key-value cache approaches to fine-tune the CLIP model, enhancing its adaptability to specific target datasets.
As an example, CLIP-Adapter incorporated a feature adapter to refine the CLIP model, enabling it to learn new features while preserving a straightforward architecture~\citep{gao2021clip}. Conversely, Tip-Adapter utilized a non-parametric key-value cache approach for training the adapter. This method bypasses backpropagation and enhances the model's adaptability to the target dataset~\citep{zhang2021tip}.
UPL reduces CLIP's dependence on labeled data by training an ensemble of prompt representations to enhance transfer performance without requiring target dataset labels~\citep{huang2022unsupervised}. Nevertheless, its zero-shot generalization effectiveness is significantly influenced by the quality of the prompt design.

From a different perspective, Shu \textit{et al}.~\citep{shu2022test} introduced the concept of test-time prompt tuning (TPT) by generating varied augmented perspectives of individual test samples, which can be effectively utilized for zero-shot generalization of the base model~\citep{shu2022test}.
Yet, the data augmentation techniques employed in TPT~\citep{shu2022test} are hindered by excessively basic variations, while relying solely on entropy-based confidence selection may not consistently ensure prediction accuracy.
To enhance TPT~\citep{shu2022test}, our conference version, DiffTPT~\citep{feng2023diverse}, proposes incorporating diffusion-based data augmentation and utilizing cosine similarity-based filtration to strike a better balance between data diversity and prediction accuracy.
In light of the versatility offered by the adapter, we have opted to swap out prompt tuning in DiffTPT for adapter. This modification allows for the comprehensive utilization of the varied enhancements in both images and text facilitated by generative models.

\subsection{Test-time Adaptation}~
Adapting machine learning models to test samples presents a more difficult and realistic scenario, as it involves the absence of training data during the inference phase~\citep{wang2020tent,sun2020test,chen2022contrastive,shanmugam2021better,zhang2023controlvideo}. This approach addresses the issues of source data being inaccessible due to privacy reasons and allows for a single training session of the model, which can then be tailored to any unforeseen test distributions~\citep{gao2022visual}.
An effective approach to creating a robust test-time objective is to decouple it from any particular training methodology. This can be achieved by either minimizing the entropy of the prediction probability distribution for the batch~\citep{wang2020tent}, or by eliminating the need for multiple test samples through the use of data augmentation techniques~\citep{zhang2021memo}.
For instance, an additional branch can be implemented to tailor the model to test samples by refining the objective during the testing phase~\citep{sun2020test,liu2021ttt++}. Wang et al.~\citep{wang2020tent} enhanced model confidence at test time by utilizing the model’s own predictions for self-adjustment, thereby minimizing the generalization error on data exhibiting distribution shifts.
An alternative method involves the explicit use of the Batch Normalization (BN) layer during test time to limit the parameters subject to optimization, thereby bolstering the model's robustness against distributional changes~\citep{schneider2020improving}.

Nonetheless, these techniques often face constraints, either due to the necessity of a substantial number of test samples for generating non-trivial solutions or due to limitations in the scalability of the model architecture.
Later research focused on leveraging large-scale pre-trained models with parameter-efficient fine-tuning~\citep{gao2022visual,zhang2022tempera}. For instance, TPT developed target-specific text prompts while keeping the main model parameters fixed. During testing, it generated various randomly augmented views and eliminated noisy augmentations that could result in inaccurate predictions by minimizing entropy~\citep{shu2022test}.
Nevertheless, entropy-based confidence selection~\citep{shu2022test} faces a limitation in effectively filtering out misclassified augmented samples that yield low-entropy predictions. 
TDA adopted a training-free dynamic adapter to enable effective test time adaptation with vision-language models~\citep{karmanov2024efficient}.
However, this method necessitates retaining each sample in the testing data stream, which is not ideal for test-time training.
To address this issue, we propose a filtration method based on cosine similarity between the logits of the original test data and the augmented images and texts (see Fig.~\ref{fig:1}).
This approach aims to ensure that the augmented samples (including image and text) maintain consistent class semantics (\ie, \textit{prediction fidelity}) while introducing \textit{diverse} information.

\subsection{Image and Text Augmentation}
Training models with synthetic images are gaining popularity and undergoing rapid development. 
In contrast to standard data augmentation methods, such as image manipulation~\citep{shorten2019survey}, image erasing~\citep{zhong2020random}, and image mixup~\citep{zhang2020does,hendrycks2019augmix}, image synthesis offers higher flexibility as these methods augment datasets with pre-defined transformations and cannot provide images with highly diverse content. Early image generation methods, including VAEs~\citep{kingma2013auto} and GANs~\citep{goodfellow2020generative}, initially provided promising generated images~\citep{brock2018large}, and have been widely applied to various vision tasks.
In the latest advancements, there has been notable progress in the creation of diffusion models, aimed at producing images of superior quality with enhanced photo-realistic features compared to earlier image generation techniques~\citep{ho2020denoising, nichol2021improved, saharia2022photorealistic, ramesh2022hierarchical, zhang2022motiondiffuse}.
New studies have illustrated the remarkable effectiveness of diffusion generative models in various applications. For instance, utilizing the latent space of powerful pretrained autoencoders has shown success in high-resolution image synthesis~\citep{rombach2022high}, improving text-guided image synthesis~\citep{nichol2021glide, ramesh2022hierarchical}, establishing a diffusion-based prior for few-shot conditional image generation~\citep{sinha2021d2c}, and implementing a probabilistic model for point cloud generation~\citep{ho2022imagen}.
The findings from these studies inspire us to enhance test data directly by integrating \textit{diverse information while maintaining consistent semantics} using the diffusion model and enhancing the performance of test-time optimization.

Another line of approaches to data augmentation involves text augmentation. Large-scale pre-trained models not only generate images with richer visual appearance variations but also provide diverse textual representations through models like GPT and other large language models (LLMs).
For example, recent research has utilized GPT-3.5 and GPT-4 to enhance data and has contrasted these approaches with conventional cutting-edge methodologies for NLP augmentation~\citep{piedboeuf2023chatgpt, ubani2023zeroshotdataaug}. The findings from these studies indicated that the application of large language models (LLMs) as data amplifiers surpassed older methodologies. This was evident particularly in tasks like rephrasing existing texts~\citep{dai2023auggpt} and creating new textual content in zero-shot scenarios using specific cues~\citep{ubani2023zeroshotdataaug}.
The \textit{diverse textual representation generation} capabilities of LLMs also offer new perspectives for test-time optimization.

\begin{figure*}[t]
	\begin{center}
		\includegraphics[width=\linewidth]{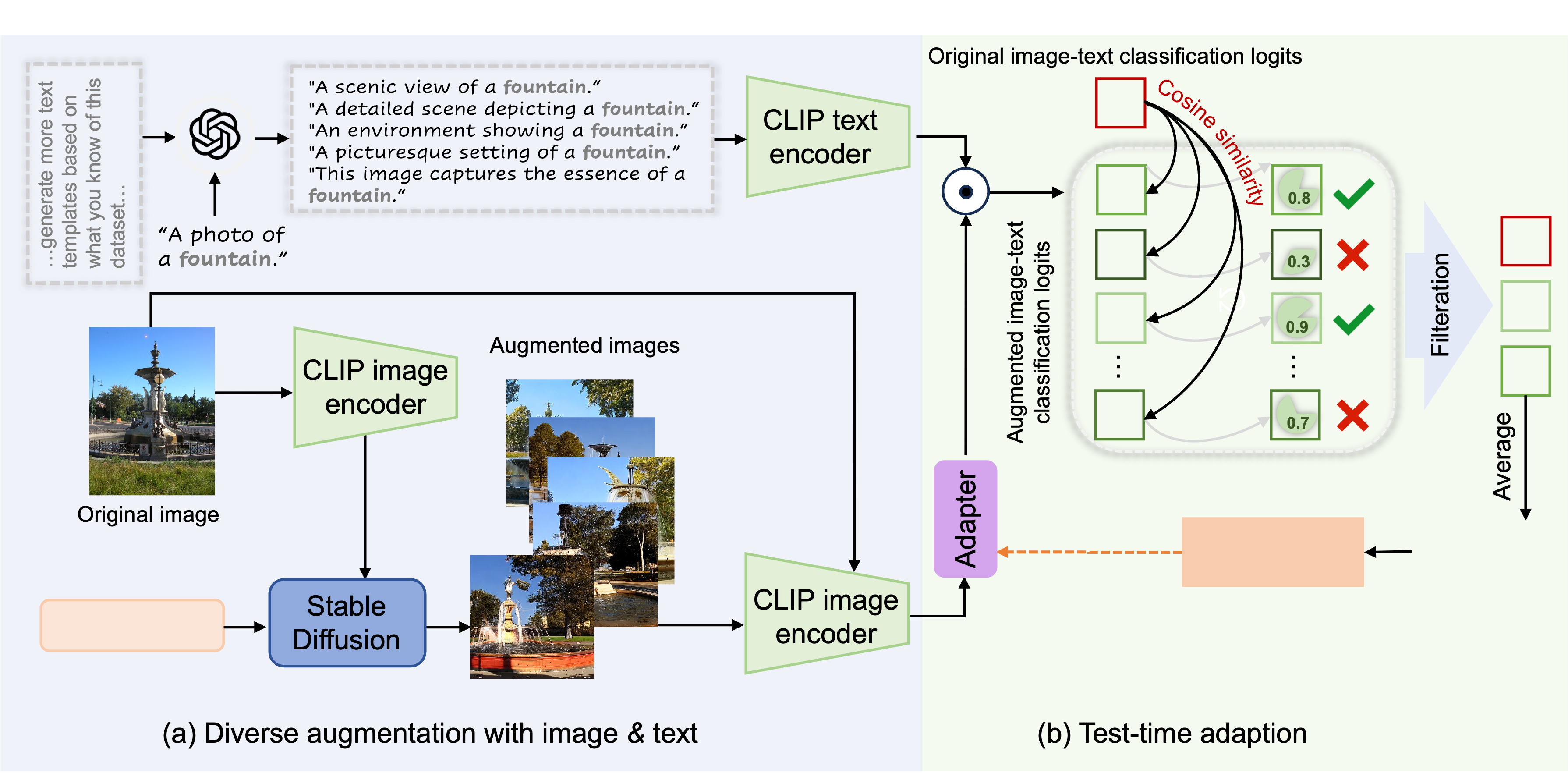}
        \put(-478,44){ \small$\mathbf{n} \sim \mathcal{N}(0, I)$}
        \put(-44,68){ \small$\tilde{\boldsymbol{p}}_{\boldsymbol{a}} (y|x)$}
        \put(-120,68){ \small $\mathop{\min}\limits 
        _{\boldsymbol{a}} \mathbf{H}\left( \tilde{\boldsymbol{p}}_{\boldsymbol{a}}\right)$}
	\end{center}
	\captionsetup{font=small}
	\caption{\small{\textbf{Overview} of our proposed \textbf{$\text{IT}^{3}\text{A}$}. First, \textbf{(a)} we utilize pre-trained generative models, \ie, diffusion and GPT-4, to generate images and text data with \textit{richer visual appearance variations and styles}. These are then randomly combined into different image-text pairs. Then, \textbf{(b)} we apply cosine similarity-based filtering on the classification logits of the augmented image text pairs generated for a single test sample against their corresponding real test sample. This helps \textit{remove spurious augmentations} and \textit{inaccurate image-text pairs}, allowing our method to \textit{balance diversity and fidelity}.}}
	\label{fig:2}
\end{figure*}

\section{Methodology}\label{sec:method}
\subsection{Approach Overview}\label{sec:moti}
Although the augmentation technique presented in the study by Shu et al.~\citep{shu2022test} has showcased notable successes in the realm of TPT, the effectiveness of this approach prominently relies on the range of diversity exhibited in the augmented images.
Given that augmented perspectives frequently exhibit akin object and background visual compositions as the original test dataset, the model confronts overly simplistic alterations within the test ensemble, potentially instigating prompt overfitting.
Furthermore, Shu \textit{et al}.~\citep{shu2022test} implement an entropy-driven confidence selection mechanism to discard augmented views displaying high entropy predictions. Essentially, the bulk of retained augmented visuals depict cropped variants of objects sourced from the initial test image (refer to Fig.~\ref{fig:1} (a), augmented views).
Consequently, the augmentation techniques outlined in~\citep{shu2022test} give rise to subtle modifications in the augmented visuals, ultimately curbing the adaptability of learned textual prompts~\citep{bansal2023leaving}.

In our conference version DiffTPT~\citep{feng2023diverse}, we employ a diffusion model on each test sample to generate diverse novel images, thereby capturing natural variations in appearance while retaining key semantics, effectively circumventing this issue. Consequently, diffusion-based data augmentation not only increases the quantity of original test samples but also maintains semantic consistency amidst distribution shifts. Furthermore, DiffTPT introduces cosine similarity-based filtering to eliminate potentially false enhancements that stable diffusion may introduce, preventing erroneous predictions.

However, we noticed that the diversity of a single modality is limited, even with powerful visual generative models. As such, we propose augmenting the original text template ``\texttt{a photo of a}'' while enhancing the visual features, refer to Fig.~\ref{fig:2} (a). To fully utilize the diverse augmentations from generative models in both images and text, we have replaced prompt tuning in DiffTPT with an adapter. Then, we have shifted the cosine similarity computation from the original test images to the logits of the original test data with augmented images and texts, thereby filtering out low-quality generated data, \ie, augmented images and text, (refer to Fig.~\ref{fig:2} (a)). Next, we will delve into the details of test time optimization based on adapters, data augmentation for images/text, and cosine similarity-based filtering across multi-modalities.

\subsection{Test-time Adaptation}\label{sec:tpt}
{
Pre-trained vision-language models such as CLIP are structured with dual encoders, including the image encoder $f(\cdot)$ and the text encoder $g(\cdot)$, offering a wealth of information for diverse downstream applications. In zero-shot classification scenarios, the prediction probabilities can be acquired by
\begin{equation}
{p}(y_i \mid \mathbf{x})=\frac{\exp \left(\cos \left(\boldsymbol{w}_{\boldsymbol{i}}, \boldsymbol{e}\right) / \tau\right)}{\sum_{j=1}^K \exp \left(\cos \left(\boldsymbol{w}_{\boldsymbol{j}}, \boldsymbol{e}\right) / \tau\right)},
\end{equation}
where the image features denoted as $\boldsymbol{e}$ generated by $f(\cdot)$ for the image $\mathbf{x}$, which in conjunction with the corresponding text feature $\boldsymbol{w}_{{i}}$ are employed to calculate the cosine similarity $\cos (\boldsymbol{w}_{{i}}, \boldsymbol{e})$ pertaining to class $i$ out of $K$ classes.
Furthermore, $\tau$ represents the temperature parameter.
Different from the TPT~\citep{shu2022test} and DiffTPT~\citep{feng2023diverse} which learn prompt embeddings, we alternatively tune the adapter to enable the use multiple augmented text templates other than initializing with single text template ``\texttt{a photo of a}''.
As such, we used CLIP-Adapter~\citep{gao2024clip} $ Adp $ for better alignment of the augmented image-text pairs proposed by generative models, hence we have
\begin{equation}
{p}(y_i \mid \mathbf{x})=\frac{\exp \left(\cos \left(\boldsymbol{w}_{\boldsymbol{i}}, \boldsymbol{e}_{\texttt{adp}}\right) / \tau\right)}{\sum_{j=1}^K \exp \left(\cos \left(\boldsymbol{w}_{\boldsymbol{j}}, \boldsymbol{e}_{\texttt{adp}}\right) / \tau\right)},
\end{equation}
where $\boldsymbol{e}_{\texttt{adp}}=\beta * Adp\left(\boldsymbol{e}\right) + (1-\beta)*\boldsymbol{e}$ , $Adp(\boldsymbol{e})$ denoting the adapted features generated by CLIP-Adapter~\citep{gao2024clip} module. $\beta$ is the coefficient in CLIP-Adapter~\citep{gao2024clip} for mixing original and adapted features.
}

{
However, as basic models typically need to generalize to out-of-distribution samples, improvements are needed in the zero-shot generalization performance of CLIP. 
TPT~\citep{shu2022test} and DiffTPT~\citep{feng2023diverse} both consider prompt tuning at test time, as it allows for modifying the context of class names to adapt to new test data samples.
However, given that different modalities can offer a more diverse knowledge, our proposed $\text{IT}^{3}\text{A}$
involves multiple text templates, breaking the restriction of using one template ``\texttt{a photo of a}'' only.
Here, we need to optimize the adapter based on a single test sample $\mathbf{x}_\texttt{test}\in \mathbb{R}^{C \times H \times W}$ during the testing phase. Formally, we have
\begin{equation}
{\boldsymbol{a}^*}=\arg \min _{\boldsymbol{a}} \mathcal{L}\left(\mathcal{F}, \boldsymbol{a}, \mathbf{x}_{\texttt{test}}\right),
\vspace{-5pt}
\end{equation}
where $\mathcal{F}$ is the CLIP model consist of an image encoder $f(\cdot)$ and a text encoder $g(\cdot)$, 
$\boldsymbol{a}^*$ denotes learnable adapter parameters of the CLIP-Adapter module.
$\mathcal{L}$ indicates the cross-entropy loss in the classification task.}

{
To ensure the efficacy of adaptation during testing, we employ a combination of various augmented pairs, \ie, image and text (\ie, $N*M$ in total), alongside a mechanism for selecting different confidence. This can be formulated as:}
\begin{equation}
\begin{aligned}
{\boldsymbol{a}^*}\!=\arg\min _{\boldsymbol{a}}-\sum_{i=1}^K \tilde{p}_{\boldsymbol{a}}\left(y_i \!\mid \!\mathbf{x}_{\texttt {test\!}}\right) \log \tilde{p}_{\boldsymbol{a}}\left(y_i \!\mid \!\mathbf{x}_{\texttt {test}}\!\right), 
\end{aligned}
\label{tpt0}
\end{equation}
\begin{equation}
\tilde{\boldsymbol{p}}_{\boldsymbol{a}}\left(y_i \!\mid \!\mathbf{x}_{\texttt{test}}\right)=\frac{1}{\rho_H NM} \sum_{n,m=1}^{N,M} \!\mathbbm{1}\!\left[\mathbf{H}_{n,m}\right] \boldsymbol{p}_{\boldsymbol{a}}\left(y_i \!\mid \!\mathcal{M}(\cdot)\right), 
\label{tpt}
\end{equation}
{where $K$ indicates the number of classes. 
%
$\mathbf{H}_{n,m}$ is a mask representing $\boldsymbol{p}_{\boldsymbol{a}}\left(y_i \mid \mathcal{M}(\cdot)\right) \leq \tau$ for filtering predictions by selection confidence levels in terms of self-entropy.
$\boldsymbol{p}_{\boldsymbol{a}}\left(y_i \mid \mathcal{M}(\cdot)\right)$ represents the class probabilities for the $n,m$-th augmented pair $\mathcal{M}(\cdot)$ 
of original pair $\left(\mathbf{x}_\texttt{test}, \mathbf{t}_\texttt{test}\right)$ from both the vision and language generative models with adapter model parameters $\boldsymbol{a}$. 
$\mathbf{t}_\texttt{text}$ is the text description for image $\mathbf{x}_\texttt{test}$ and category with label $y_i$.
%
The threshold $\tau$ determines the selection of confidence levels that lead to a $\rho_H$ fraction of all $N*M$ augmented pairs. 
%
}

\subsection{Diverse Data Augmentation}\label{sec:difftpt}
\subsubsection{Diffusion-based Diverse Image Augmentation}
{
Diffusion-driven image augmentation aims to create a range of varied and enriching augmented images.
As illustrated in Fig.~\ref{fig:2} (a), we begin by extracting latent features $z_0$ from the pre-trained CLIP encoder $f(\mathbf{x}_\text{test})$ of a given test image $\mathbf{x}_\text{test}$, followed by employing stable diffusion as the decoder to generate diverse augmented images.
Here, we utilize the Stable Diffusion-V2 model as the generative framework, enabling the creation of a novel image $\mathcal G(g(\mathbf{t}), \mathbf{n})$ based on textual descriptions $\mathbf{t}$.
$\mathbf{n} \sim \mathcal{N}(0, I)$ indicates the sampled noise.
Given the absence of labels during test-time tuning, we opt to substitute $g(\mathbf{t})$ with the image encoder from the CLIP model, denoted as $f(\mathbf{x}_\texttt{test})$.
Consequently, the synthetic image can be produced by utilizing
\begin{equation}
\mathcal D_n(\mathbf{x}_\texttt{test}) = \mathcal G(f(\mathbf{x}_\texttt{test}), \mathbf{n}_n),
\vspace{-5pt}
\end{equation}
where $n$-th augmented image is represented as $\mathcal D_n(\mathbf{x}_\text{test})$. 
The alignment capability of CLIP in associating images with text leads to the efficacy of diffusion-driven data augmentation in creating a varied set of augmented images. 
Fig.~\ref{fig:3.4} depicts our diverse and informative augmentations.
}

{
By incorporating the augmented images ${\mathcal D_n(\mathbf{x}_\texttt{test})}$, modifications to the paired augmentation $\mathcal{M}\left(\cdot\right)$ as outlined in Eq.~(\ref{tpt}) can be achieved through adaptation:
\begin{equation}
\mathcal{M}\left(\cdot\right)=\left(\mathcal{D}_{n}\left(\mathbf{x}_{\texttt {test}}\right), \mathcal{T}_{m}\left(\mathbf{t}_{\texttt{test}}\right)\right)
\end{equation}
where $\mathcal{T}_{m}\left(\mathbf{t}_{\texttt{test}}\right)$ denotes the $m$-th augmented text.
Please note that the augmented data from different views in TPT~\citep{shu2022test} are both incorporated with diffusion-based ones to take advantage of their complementary merits. 
IAs a consequence, we can still use Eq.~(\ref{tpt0}) and Eq.~(\ref{tpt}) for test time adaptation.  
}

\subsubsection{{LLM-based Diverse Text Augmentation}}
{For text augmentation, we employ generative Large Language Models (LLMs), to obtain diverse text templates. As is shown in Fig.~\ref{fig:2} (a), from a given single text template $\mathbf{t}_\texttt{test}$, we instruct the LLM to generate augmented text templates. Specifically, we employ GPT-4~\citep{achiam2023gpt} as the generative language model, which can generate new template $\mathcal{K}(\mathbf{t}_\texttt{test}, \mathbf{s})$, with natural language LLM instruction $\mathbf{s}$. Thus, the augmented template can then be generated with
\begin{equation}
\mathcal{T}_m(\mathbf{t}_{\texttt{test}})=\mathcal{K}(\mathbf{t}_\texttt{test}, \mathbf{s}),
\end{equation}
where $\mathcal{T}_m(\mathbf{t}_{\texttt{test}})$ denotes the $m$-th augmented template. Below is an example of the $\mathbf{s}$.}

\begin{figure}[H]
	\begin{center}
		\includegraphics[width=\linewidth]{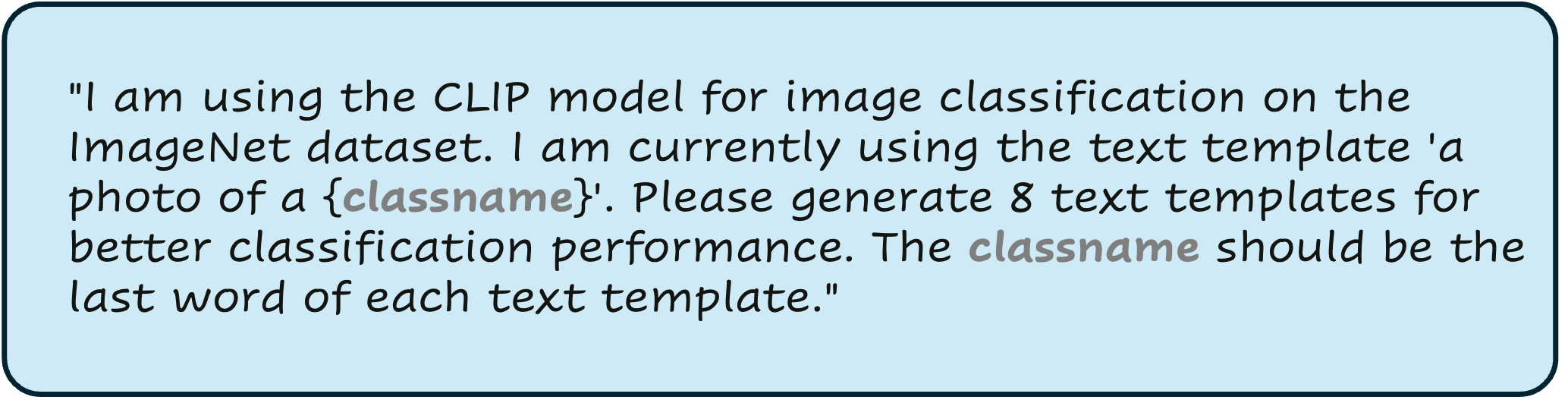}
	\end{center}
	\captionsetup{font=small}
\end{figure}

\subsection{Filtration with Cosine Similarity}\label{sec:cosine}
\begin{figure*}[t]
	\begin{center}
		\includegraphics[width=\linewidth]{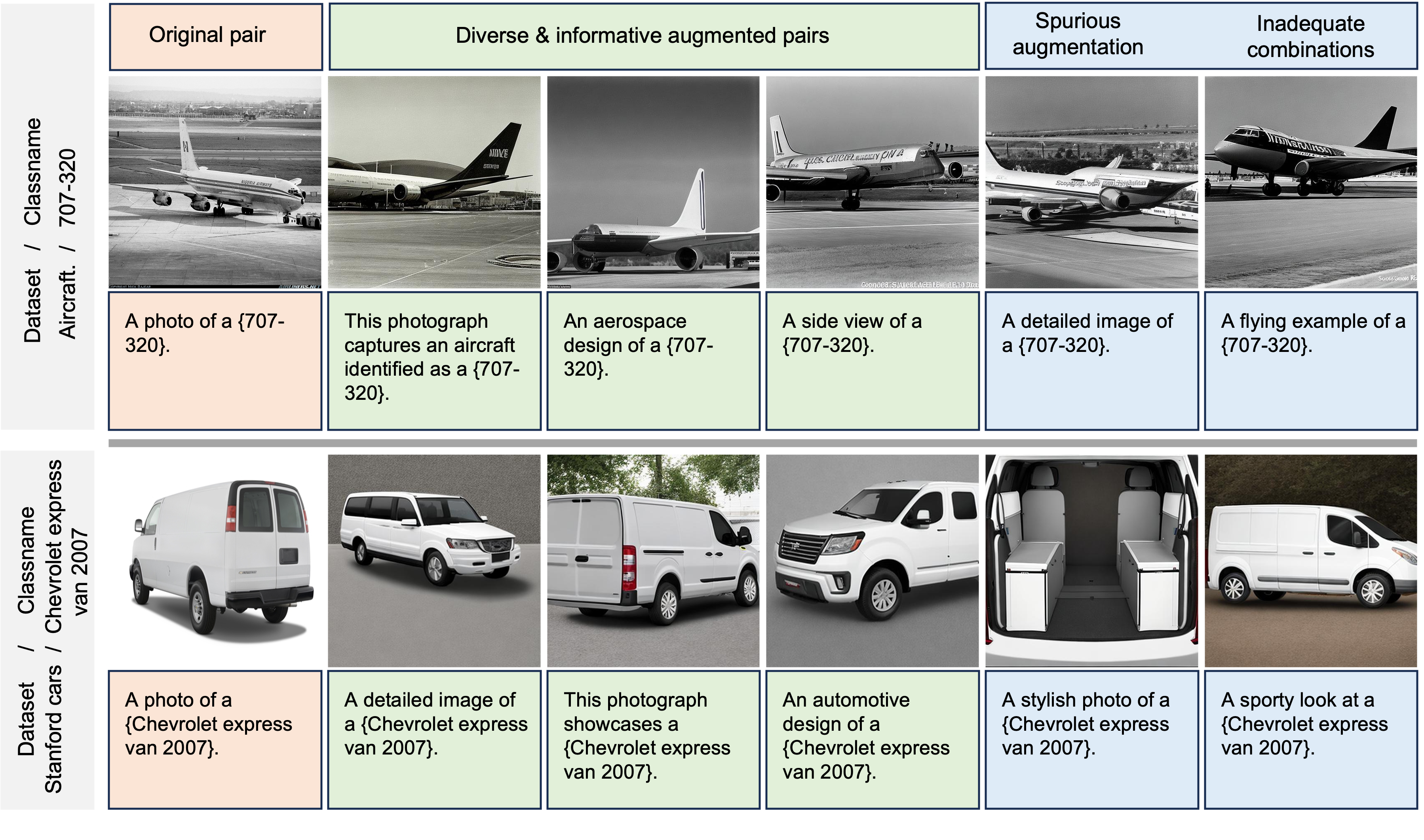}
	\end{center}
	\captionsetup{font=small}
    	\caption{{Examples of the \textbf{original pairs} ({\color{Melon}orange box}) from a single test data include \textbf{diverse and informative augmented pairs} ({\color{DarkSeaGreen1}green box}), as well as \textbf{spurious augmentations} and \textbf{inadequate combinations} ({\color{LightBlue1}blue box}). Spurious augmentations and inadequate combinations are filtered using the cosine similarity of the predicted logits from the image-text pairs.}}
	\label{fig:3.4}
\end{figure*}

Although data augmentation based on generative models is effective in producing a variety of augmented data, it may introduce some spurious augmentations (see Fig.~\ref{fig:3.4}), resulting in low data fidelity and degraded performance during learning.
These false augmentations stem partly from spurious augmentation generated by the diffusion model itself and partly from inadequate combinations of text and generated images (see the blue box in Fig.~\ref{fig:3.4}).
Therefore, it is essential to balance the diversity of augmented data with the fidelity of predictions.

To achieve this, we use cosine similarity-based filtering to remove the above false augmentations, \ie, low-quality images generated by the diffusion model, and inadequate combinations of text and generated images.
In specific, we calculate the cosine similarity between the classification logits of the test sample pair $(\mathbf{x}_{\texttt{test}}, \mathbf{t}_{\texttt{test}})$ and each augmented image-text pair $\mathcal{M}\left(\cdot\right) = (\mathcal{D}_n(\mathbf{x}_{\texttt{test}}), \mathcal{T}_m(\mathbf{t}_\texttt{test}))$.
We then introduce a mask $\mathcal{C}$ to identify augmented data with a similarity exceeding $\varepsilon$. Formally, $\mathcal{C}_{n,m} = \cos(\mathbf{l}_0, \mathbf{l}_{n,m}) > \varepsilon$, where $\mathbf{l}_0$ and $\mathbf{l}_{n,m}$ represent the classification logits of the test sample pair $(\mathbf{x}_{\text{test}}, \mathbf{t}_{\text{test}})$ and each augmented image-text pair $\mathcal{M}(\cdot)$, respectively.
We note that $\varepsilon$ is the threshold parameter that leads to a $\rho_C$ percentage of the augmented image-text pairs.
Formally, the test-time adaptation in Eq.~(\ref{tpt}) can be further modified as
\begin{equation}
\tilde{\boldsymbol{p}}_{\boldsymbol{a}}\left(y_i \!\mid \!\mathbf{x}_{\texttt{test}}\right)= \frac{1}{Z}\sum_{n,m=1}^{N,M} \!\mathbbm{1}\!\left[\mathbf{H}_{n,m}\right] \cdot \!\mathbbm{1}\!\left[\mathcal{C}_{n,m}\right] 
\boldsymbol{p}_{\boldsymbol{a}}\left(y_i \!\mid \!\mathcal{M}\left(\cdot\right)\right),
\vspace{-2pt}
\label{difftpt2}
\end{equation}
{where $\frac{1}{Z} = \frac{1}{\rho_H \rho_C NM}$.
As a result, we can generate a substantial number of augmented samples with greater diverse data, while retaining essential semantics to refine the adapter during test-time.
}

\section{Experiments}
\subsection{Experimental Setup}
\noindent{\textbf{Implementation Details.}} 
{
Our experiments are conducted with $32$GB NVIDIA Tesla V$100$ GPUs and $40$GB NVIDIA A100 GPUs, each run requiring one GPU.
For CLIP-Adapter~\citep{gao2024clip}, the initial weights are randomly initialized, and the adapter model is fine-tuned based on a single test image. By default, the dimensionality reduction for the adapter is set to $4$.
DiffTPT~\citep{shu2022test} enhances each test image to create variations via Stable Diffusion-V2 and through diverse augment views~\citep{shu2022test}. The number of variations is set to $7$ for both stable diffusion and augment view.
For our method, we generate $7$ new images and further enhance the image-text pairs with $7$ different text templates generated by GPT-4.
The adapter undergoes optimization over $4$ steps during the test phase using the AdamW optimizer, with the initial learning rate, $\rho_{H}$, and $\rho_{C}$ set to $0.005$, $0.3$, and $0.8$, respectively.
}

\noindent{\textbf{Datasets.}} 
{
We use two $\mathcal{S}$cenarios to evaluate our proposed method, \ie, $\mathcal{S}_1$: \texttt{Natural Distribution Shifts} and $\mathcal{S}_2$: \texttt{Cross-Dataset Generalization}.
For $\mathcal{S}_1$, following~\citep{shu2022test}, we use four out-of-distribution (OOD) datasets including \textbf{ImageNet}~\citep{deng2009imagenet}, \ie, \textbf{ImageNet-V2}~\citep{recht2019imagenet}, \textbf{ImageNet-A}~\citep{hendrycks2021natural}, \textbf{ImageNet-R}~\citep{hendrycks2021many}, and \textbf{ImageNet-Sketch}~\citep{wang2019learning}.
These datasets vary in image style and data domains, allowing us to evaluate the robustness of our method against natural distribution shifts.
For $\mathcal{S}_2$, we utilize $10$ diverse datasets covering various species of plants and animals, scenes, textures, food, transportation, human actions, satellite images, and general objects: \textbf{Flower102}~\citep{nilsback2008automated}, \textbf{OxfordPets}~\citep{parkhi2012cats}, \textbf{SUN397}~\citep{xiao2010sun}, \textbf{DTD}~\citep{cimpoi2014describing}, \textbf{Food101}~\citep{bossard2014food}, \textbf{StanfordCars}~\citep{krause20133d}, \textbf{Aircraft}~\citep{maji2013fine}, \textbf{UCF101}~\citep{soomro2012ucf101}, \textbf{EuroSAT}~\citep{helber2019eurosat}, and \textbf{Caltech101}~\citep{fei2004learning}. 
To explore cross-dataset generalization, ImageNet serves as the source dataset, while the other $10$ datasets are used as target datasets for evaluation. In our experiments, we randomly select $1,000$ test images from all classes to evaluate each method.
}

\noindent{\textbf{Baselines.}}
{
To assess our proposed method, we employ three groups of methodologies: \textbf{a)} TPT~\citep{shu2022test}, which is a state-of-the-art test-time prompt tuning technique optimized using multiple augmented views, \textbf{b)} traditional PEFT methods for CLIP, specifically CoOp~\citep{zhou2022learning} the few-shot prompt tuning baseline that adjusts a fixed prompt for each downstream dataset, CoCoOp~\citep{zhou2022conditional} the enhanced few-shot prompt tuning baseline that creates input-conditional prompts via a lightweight neural network, and CLIP-Adapter~\citep{gao2024clip}, a flexible method which help model adapt to new dataset at feature level, and \textbf{c)} zero-shot CLIP, using the default prompt ``\texttt{a photo of a}''. Adhering to the procedures of previous works~\citep{zhou2022learning,zhou2022conditional,shu2022test,gao2024clip}, all baselines are trained on ImageNet with $16$-shot examples, $4$ learnable prompt tokens for CoOp/CoCoOp, $2$-layer linear adapter for CLIP-Adapter~\citep{gao2024clip} and subsequently tested on OOD benchmarks.
In TPT and DiffTPT, default template ``\texttt{a photo of a}'' and CoOp/CoCoOp pretrained weights are used for initialization. 
We note that such methods to initialize learnable prompts restrict the use of multiple augment templates.
$\text{IT}^{3}\text{A}$ adopts CLIP-Adapter~\citep{gao2024clip} as a more flexible backbone.
}

\begin{table*}[t]
\renewcommand{\arraystretch}{1.3}
	\caption{{\textbf{Top 1 accuracy} $\%$ of state-of-the-art baselines under Scenario 1 ($\mathcal{S}_1$). \textbf{ImageNet-Sk.} denotes the ImageNet-Sketch dataset, while \textbf{OOD Avg.} represents the average performance across out-of-distribution datasets. The abbreviation {\cellcolor{mygray}{$bs.$}} signifies the baseline for each group, \ie, CLIP-RN50 / CLIP-ViT-B-16, CoOp, CoCoOp, and CLIP-Adapter. Arrows (${\color{ForestGreen}\uparrow}$ and ${\color{red}\downarrow}$) indicate enhancements and reductions compared to the baseline. For comprehensive analyses, refer to Sec.~\ref{sec:acc}.}}
	\label{tab:1}
        
	\fontsize{8.5}{8.5}\selectfont
	\centering
	\begin{tabular}{l cc cc cc cc cc cc cc}
\toprule

\textbf{Method}
&\multicolumn{2}{c}{\textbf{~~ImageNet~~}}
&\multicolumn{2}{c}{\textbf{~~ImageNet-A~~}}
&\multicolumn{2}{c}{\textbf{~ImageNet-V2~}} 
&\multicolumn{2}{c}{\textbf{~~ImageNet-R~~}}
&\multicolumn{2}{c}{\textbf{~ImageNet-Sk.~}} 
&\multicolumn{2}{c}{\textbf{~~~~Average~~~~}} 
&\multicolumn{2}{c}{\textbf{~~OOD Avg.~~}}\\
\cmidrule(r){1-1}  \cmidrule{2-3} \cmidrule(lr){4-5} \cmidrule(lr){6-7} \cmidrule(lr){8-9}
\cmidrule(lr){10-11} \cmidrule(lr){12-13} \cmidrule(lr){14-15}

{{\cellcolor{mygray}CLIP-RN50}} 
&\multicolumn{2}{|c}{{\cellcolor{mygray}{$56.70$\stdvno{$bs.$}}}}
&\multicolumn{2}{c}{{\cellcolor{mygray}{$23.80$\stdvno{$bs.$}} }}
&\multicolumn{2}{c}{{\cellcolor{mygray}{$50.20$\stdvno{$bs.$}} }}
&\multicolumn{2}{c}{{\cellcolor{mygray}{$54.40$\stdvno{$bs.$}}}}
&\multicolumn{2}{c}{{\cellcolor{mygray}{$33.70$\stdvno{$bs.$}} }}
&\multicolumn{2}{|c}{{\cellcolor{mygray}{$43.76$\stdvno{$bs.$}} }}
&\multicolumn{2}{|c}{{\cellcolor{mygray}{$40.53$\stdvno{$bs.$}}}} \\

{\cellcolor{mypink2}TPT }
&\multicolumn{2}{|c}{{\cellcolor{mypink2}$56.80$\stdvu{$0.10$} }}
&\multicolumn{2}{c}{{\cellcolor{mypink2}$23.80$\stdvu{$0.00$} } }
&\multicolumn{2}{c}{{\cellcolor{mypink2}$50.30$\stdvu{$0.10$}} }
&\multicolumn{2}{c}{{\cellcolor{mypink2}$54.30$\stdvd{$0.10$}}}
&\multicolumn{2}{c}{{\cellcolor{mypink2}$33.60$\stdvd{$0.10$}} }
&\multicolumn{2}{|c}{{\cellcolor{mypink2}$43.76$\stdvu{$0.00$}} }
&\multicolumn{2}{|c}{{\cellcolor{mypink2}$40.50$\stdvd{$0.03$}}} \\

{{\cellcolor{mypink2}{DiffTPT}}} 
&\multicolumn{2}{|c}{{\cellcolor{mypink2}${58.00}$\stdvu{${{1.30}}$}}}
&\multicolumn{2}{c}{{\cellcolor{mypink2}${31.40}$\stdvu{${{7.60}}$}}}
&\multicolumn{2}{c}{{\cellcolor{mypink2}${51.80}$\stdvu{${{1.60}}$}}}
&\multicolumn{2}{c}{{\cellcolor{mypink2}${56.50}$\stdvu{${{2.10}}$}}}
&\multicolumn{2}{c}{{\cellcolor{mypink2}${35.80}$\stdvu{${{2.10}}$}}}
&\multicolumn{2}{|c}{{\cellcolor{mypink2}${46.70}$\stdvu{${{2.94}}$}}} 
&\multicolumn{2}{|c}{{\cellcolor{mypink2}${43.88}$\stdvu{${{3.35}}$}}} \\

{{\cellcolor{mypink2}\textbf{$\text{IT}^{3}\text{A}$}}} 
&\multicolumn{2}{|c}{{\cellcolor{mypink2}$\textbf{59.30}$\stdvu{${{2.60}}$}}}
&\multicolumn{2}{c}{{\cellcolor{mypink2}$\textbf{{33.90}}$\stdvu{${{10.10}}$}}}
&\multicolumn{2}{c}{{\cellcolor{mypink2}$\textbf{54.30}$\stdvu{${{4.10}}$}}}
&\multicolumn{2}{c}{{\cellcolor{mypink2}$\textbf{59.90}$\stdvu{${{5.50}}$}}}
&\multicolumn{2}{c}{{\cellcolor{mypink2}$\textbf{38.90}$\stdvu{${{5.20}}$}}}
&\multicolumn{2}{|c}{{\cellcolor{mypink2}$\textbf{49.26}$\stdvu{${{5.50}}$}}} 
&\multicolumn{2}{|c}{{\cellcolor{mypink2}$\textbf{46.75}$\stdvu{${{6.22}}$}}} \\

\cmidrule(r){1-15}

{\cellcolor{mypink1}CoOp }
&\multicolumn{2}{|c}{{\cellcolor{mypink1}$62.00$\stdvno{$bs.$}}}
&\multicolumn{2}{c}{{\cellcolor{mypink1}$25.00$\stdvno{$bs.$}} }
&\multicolumn{2}{c}{{\cellcolor{mypink1}$54.60$\stdvno{$bs.$}} }
&\multicolumn{2}{c}{{\cellcolor{mypink1}$54.70$\stdvno{$bs.$}}}
&\multicolumn{2}{c}{{\cellcolor{mypink1}$36.00$\stdvno{$bs.$}} }
&\multicolumn{2}{|c}{{\cellcolor{mypink1}$46.46$\stdvno{$bs.$}} }
&\multicolumn{2}{|c}{{\cellcolor{mypink1}$42.58$\stdvno{$bs.$}}} \\

{\cellcolor{mypink1}TPT\&CoOp }
&\multicolumn{2}{|c}{{\cellcolor{mypink1}$62.00$\stdvu{$0.00$} }}
&\multicolumn{2}{c}{{\cellcolor{mypink1}$25.00$\stdvu{$0.00$} } }
&\multicolumn{2}{c}{{\cellcolor{mypink1}$54.90$\stdvu{$0.30$}} }
&\multicolumn{2}{c}{{\cellcolor{mypink1}$54.90$\stdvu{$0.20$}}}
&\multicolumn{2}{c}{{\cellcolor{mypink1}$36.40$\stdvu{$0.40$}} }
&\multicolumn{2}{|c}{{\cellcolor{mypink1}$46.64$\stdvu{$0.18$}} }
&\multicolumn{2}{|c}{{\cellcolor{mypink1}$42.80$\stdvu{$0.22$}}} \\

{{\cellcolor{mypink1}{DiffTPT\&CoOp}}} 
&\multicolumn{2}{|c}{{\cellcolor{mypink1}$\underline{63.00}$\stdvu{${{1.00}}$}}}
&\multicolumn{2}{c}{{\cellcolor{mypink1}$\underline{33.70}$\stdvu{${{8.70}}$}}}
&\multicolumn{2}{c}{{\cellcolor{mypink1}$\underline{55.70}$\stdvu{${{1.10}}$}}}
&\multicolumn{2}{c}{{\cellcolor{mypink1}$\underline{57.60}$\stdvu{${{2.90}}$}}}
&\multicolumn{2}{c}{{\cellcolor{mypink1}${34.60}$\stdvd{${{1.40}}$}}}
&\multicolumn{2}{|c}{{\cellcolor{mypink1}$\underline{48.92}$\stdvu{${{2.46}}$}}} 
&\multicolumn{2}{|c}{{\cellcolor{mypink1}$\underline{45.40}$\stdvu{${{2.82}}$}}} \\

{\cellcolor{mypink}CoCoOp }
&\multicolumn{2}{|c}{{\cellcolor{mypink}$58.20$\stdvno{$bs.$} }}
&\multicolumn{2}{c}{{\cellcolor{mypink}$26.50$\stdvno{$bs.$} } }
&\multicolumn{2}{c}{{\cellcolor{mypink}$53.10$\stdvno{$bs.$}} }
&\multicolumn{2}{c}{{\cellcolor{mypink}$55.90$\stdvno{$bs.$}}}
&\multicolumn{2}{c}{{\cellcolor{mypink}$35.90$\stdvno{$bs.$}} }
&\multicolumn{2}{|c}{{\cellcolor{mypink}$45.92$\stdvno{$bs.$}} }
&\multicolumn{2}{|c}{{\cellcolor{mypink}$42.85$\stdvno{$bs.$}}} \\

{\cellcolor{mypink}TPT\&CoCoOp }
&\multicolumn{2}{|c}{{\cellcolor{mypink}$58.20$\stdvu{$0.00$} }}
&\multicolumn{2}{c}{{\cellcolor{mypink}$26.50$\stdvu{$0.00$} } }
&\multicolumn{2}{c}{{\cellcolor{mypink}$53.20$\stdvu{$0.10$}} }
&\multicolumn{2}{c}{{\cellcolor{mypink}$55.90$\stdvu{$0.00$}}}
&\multicolumn{2}{c}{{\cellcolor{mypink}$35.80$\stdvd{$0.10$}} }
&\multicolumn{2}{|c}{{\cellcolor{mypink}$45.92$\stdvu{$0.00$}} }
&\multicolumn{2}{|c}{{\cellcolor{mypink}$42.85$\stdvu{$0.00$}} }\\

{{\cellcolor{mypink}{DiffTPT\&CoCoOp}}} 
&\multicolumn{2}{|c}{{\cellcolor{mypink}${58.30}$\stdvu{${{0.10}}$}}}
&\multicolumn{2}{c}{{\cellcolor{mypink}${26.30}$\stdvd{${{0.20}}$}}}
&\multicolumn{2}{c}{{\cellcolor{mypink}${53.30}$\stdvu{${{0.20}}$}}}
&\multicolumn{2}{c}{{\cellcolor{mypink}${56.00}$\stdvu{${{0.10}}$}}}
&\multicolumn{2}{c}{{\cellcolor{mypink}$\underline{35.90}$\stdvu{${{0.00}}$}}}
&\multicolumn{2}{|c}{{\cellcolor{mypink}${45.94}$\stdvu{${{0.02}}$}}} 
&\multicolumn{2}{|c}{{\cellcolor{mypink}${42.88}$\stdvu{${{0.03}}$}}} \\

{{\cellcolor{mypink}{CLIP-Ap.}}} 
&\multicolumn{2}{|c}{{\cellcolor{mypink}${60.50}$\stdvno{$bs.$}}}
&\multicolumn{2}{c}{{\cellcolor{mypink}${24.10}$\stdvno{$bs.$}}}
&\multicolumn{2}{c}{{\cellcolor{mypink}${53.30}$\stdvno{$bs.$}}}
&\multicolumn{2}{c}{{\cellcolor{mypink}${54.70}$\stdvno{$bs.$}}}
&\multicolumn{2}{c}{{\cellcolor{mypink}${34.50}$\stdvno{$bs.$}}}
&\multicolumn{2}{|c}{{\cellcolor{mypink}${45.42}$\stdvno{$bs.$}}}
&\multicolumn{2}{|c}{{\cellcolor{mypink}${41.65}$\stdvno{$bs.$}}
} \\

{{\cellcolor{mypink}\textbf{$\text{IT}^{3}\text{A}$\&CLIP-Ap.}}} 
&\multicolumn{2}{|c}{{\cellcolor{mypink}$\textbf{62.40}$\stdvu{${{1.90}}$}}}
&\multicolumn{2}{c}{{\cellcolor{mypink}$\textbf{33.80}$\stdvu{${{9.70}}$}}}
&\multicolumn{2}{c}{{\cellcolor{mypink}$\textbf{56.30}$\stdvu{${{3.00}}$}}}
&\multicolumn{2}{c}{{\cellcolor{mypink}$\textbf{60.60}$\stdvu{${{5.90}}$}}}
&\multicolumn{2}{c}{{\cellcolor{mypink}$\textbf{36.60}$\stdvu{${{2.10}}$}}}
&\multicolumn{2}{|c}{{\cellcolor{mypink}$\textbf{49.94}$\stdvu{${{4.52}}$}}} 
&\multicolumn{2}{|c}{{\cellcolor{mypink}$\textbf{46.83}$\stdvu{${{5.18}}$}}} \\

\hline\hline

{{\cellcolor{mygray}{CLIP-ViT-B/16} }}
&\multicolumn{2}{|c}{{\cellcolor{mygray}{$63.60$\stdvno{$bs.$} }}}
&\multicolumn{2}{c}{{\cellcolor{mygray}{$47.20$\stdvno{$bs.$} }}} 
&\multicolumn{2}{c}{{\cellcolor{mygray}{$59.40$\stdvno{$bs.$}}}} 
&\multicolumn{2}{c}{{\cellcolor{mygray}{$72.60$\stdvno{$bs.$}}}}
&\multicolumn{2}{c}{{\cellcolor{mygray}{$46.00$\stdvno{$bs.$}} }}
&\multicolumn{2}{|c}{{\cellcolor{mygray}{$57.76$\stdvno{$bs.$}}}} 
&\multicolumn{2}{|c}{{\cellcolor{mygray}{$56.30$\stdvno{$bs.$}}}} \\

{\cellcolor{mypink2}TPT }
&\multicolumn{2}{|c}{{\cellcolor{mypink2}$63.60$\stdvu{$0.00$} }}
&\multicolumn{2}{c}{{\cellcolor{mypink2}$47.40$\stdvu{$0.20$} } }
&\multicolumn{2}{c}{{\cellcolor{mypink2}$59.50$\stdvu{$0.10$}} }
&\multicolumn{2}{c}{{\cellcolor{mypink2}$72.70$\stdvu{$0.10$}}}
&\multicolumn{2}{c}{{\cellcolor{mypink2}$45.90$\stdvd{$0.10$}} }
&\multicolumn{2}{|c}{{\cellcolor{mypink2}$57.82$\stdvu{$0.06$}} }
&\multicolumn{2}{|c}{{\cellcolor{mypink2}$56.38$\stdvu{$0.08$}}} \\

{{\cellcolor{mypink2}{DiffTPT}}} 
&\multicolumn{2}{|c}{{\cellcolor{mypink2}${64.80}$\stdvu{${{1.20}}$}}}
&\multicolumn{2}{c}{{\cellcolor{mypink2}$\textbf{54.50}$\stdvu{${{7.30}}$}}}
&\multicolumn{2}{c}{{\cellcolor{mypink2}${60.10}$\stdvu{${{0.70}}$}}}
&\multicolumn{2}{c}{{\cellcolor{mypink2}${74.30}$\stdvu{${{1.70}}$}}}
&\multicolumn{2}{c}{{\cellcolor{mypink2}${47.50}$\stdvu{${{1.50}}$}}}
&\multicolumn{2}{|c}{{\cellcolor{mypink2}${60.24}$\stdvu{${{2.48}}$}}} 
&\multicolumn{2}{|c}{{\cellcolor{mypink2}${59.10}$\stdvu{${{2.80}}$}}} \\

{{\cellcolor{mypink2}\textbf{$\text{IT}^{3}\text{A}$}}} 
&\multicolumn{2}{|c}{{\cellcolor{mypink2}$\textbf{66.00}$\stdvu{${{2.40}}$}}}
&\multicolumn{2}{c}{{\cellcolor{mypink2}${51.30}$\stdvu{${{4.10}}$}}}
&\multicolumn{2}{c}{{\cellcolor{mypink2}$\textbf{60.70}$\stdvu{${{1.30}}$}}}
&\multicolumn{2}{c}{{\cellcolor{mypink2}$\textbf{76.00}$\stdvu{${{3.40}}$}}}
&\multicolumn{2}{c}{{\cellcolor{mypink2}$\textbf{49.00}$\stdvu{${{3.00}}$}}}
&\multicolumn{2}{|c}{{\cellcolor{mypink2}$\textbf{60.60}$\stdvu{${{2.84}}$}}} 
&\multicolumn{2}{|c}{{\cellcolor{mypink2}$\textbf{59.25}$\stdvu{${{2.95}}$}}} \\

\cmidrule(r){1-15}

{\cellcolor{mypink1}CoOp }
&\multicolumn{2}{|c}{{\cellcolor{mypink1}$68.30$\stdvno{$bs.$} }}
&\multicolumn{2}{c}{{\cellcolor{mypink1}$48.10$\stdvno{$bs.$} } }
&\multicolumn{2}{c}{{\cellcolor{mypink1}$61.90$\stdvno{$bs.$}} }
&\multicolumn{2}{c}{{\cellcolor{mypink1}$70.70$\stdvno{$bs.$}}}
&\multicolumn{2}{c}{{\cellcolor{mypink1}$45.50$\stdvno{$bs.$}} }
&\multicolumn{2}{|c}{{\cellcolor{mypink1}$58.90$\stdvno{$bs.$}} }
&\multicolumn{2}{|c}{{\cellcolor{mypink1}$56.55$\stdvno{$bs.$}}} \\

{\cellcolor{mypink1}TPT\&CoOp }
&\multicolumn{2}{|c}{{\cellcolor{mypink1}$68.30$\stdvu{$0.00$} }}
&\multicolumn{2}{c}{{\cellcolor{mypink1}$48.20$\stdvu{$0.10$} } }
&\multicolumn{2}{c}{{\cellcolor{mypink1}$62.00$\stdvu{$0.10$}} }
&\multicolumn{2}{c}{{\cellcolor{mypink1}$70.70$\stdvu{$0.00$}}}
&\multicolumn{2}{c}{{\cellcolor{mypink1}$45.60$\stdvu{$0.10$}} }
&\multicolumn{2}{|c}{{\cellcolor{mypink1}$58.94$\stdvu{$0.04$}} }
&\multicolumn{2}{|c}{{\cellcolor{mypink1}$56.60$\stdvu{$0.05$}}} \\

{{\cellcolor{mypink1}{DiffTPT\&CoOp}}} 
&\multicolumn{2}{|c}{{\cellcolor{mypink1}$\underline{69.70}$\stdvu{${{1.40}}$}}}
&\multicolumn{2}{c}{{\cellcolor{mypink1}$\underline{53.00}$\stdvu{${{4.90}}$}}}
&\multicolumn{2}{c}{{\cellcolor{mypink1}${62.30}$\stdvu{${{0.40}}$}}}
&\multicolumn{2}{c}{{\cellcolor{mypink1}${72.60}$\stdvu{${{1.90}}$}}}
&\multicolumn{2}{c}{{\cellcolor{mypink1}${46.50}$\stdvu{${{1.00}}$}}}
&\multicolumn{2}{|c}{{\cellcolor{mypink1}$\underline{60.82}$\stdvu{${{1.92}}$}}} 
&\multicolumn{2}{|c}{{\cellcolor{mypink1}$\underline{58.60}$\stdvu{${{2.05}}$}}} \\

{\cellcolor{mypink}CoCoOp }
&\multicolumn{2}{|c}{{\cellcolor{mypink}$65.90$\stdvno{$bs.$} }}
&\multicolumn{2}{c}{{\cellcolor{mypink}$48.90$\stdvno{$bs.$} } }
&\multicolumn{2}{c}{{\cellcolor{mypink}$60.90$\stdvno{$bs.$}} }
&\multicolumn{2}{c}{{\cellcolor{mypink}$74.50$\stdvno{$bs.$}}}
&\multicolumn{2}{c}{{\cellcolor{mypink}$47.80$\stdvno{$bs.$}} }
&\multicolumn{2}{|c}{{\cellcolor{mypink}$59.60$\stdvno{$bs.$}} }
&\multicolumn{2}{|c}{{\cellcolor{mypink}$58.03$\stdvno{$bs.$}}} \\

{\cellcolor{mypink}TPT\&CoCoOp }
&\multicolumn{2}{|c}{{\cellcolor{mypink}$65.90$\stdvu{$0.00$}}}
&\multicolumn{2}{c}{{\cellcolor{mypink}$48.80$\stdvd{$0.10$} } }
&\multicolumn{2}{c}{{\cellcolor{mypink}$60.90$\stdvu{$0.00$}} }
&\multicolumn{2}{c}{{\cellcolor{mypink}$\underline{\textbf{74.60}}$\stdvu{$0.10$}}}
&\multicolumn{2}{c}{{\cellcolor{mypink}$47.80$\stdvu{$0.00$}} }
&\multicolumn{2}{|c}{{\cellcolor{mypink}$59.60$\stdvu{$0.00$}} }
&\multicolumn{2}{|c}{{\cellcolor{mypink}$58.03$\stdvu{$0.00$}} }\\

{{\cellcolor{mypink}{DiffTPT\&CoCoOp}}} 
&\multicolumn{2}{|c}{{\cellcolor{mypink}${66.90}$\stdvu{${{1.00}}$}}}
&\multicolumn{2}{c}{{\cellcolor{mypink}${48.70}$\stdvd{${{0.20}}$}}}
&\multicolumn{2}{c}{{\cellcolor{mypink}${61.80}$\stdvu{${{0.90}}$}}}
&\multicolumn{2}{c}{{\cellcolor{mypink}${74.50}$\stdvu{${{0.00}}$}}}
&\multicolumn{2}{c}{{\cellcolor{mypink}$\underline{49.10}$\stdvu{${{1.30}}$}}}
&\multicolumn{2}{|c}{{\cellcolor{mypink}${60.20}$\stdvu{${{0.60}}$}}} 
&\multicolumn{2}{|c}{{\cellcolor{mypink}${58.53}$\stdvu{${{0.50}}$}}} \\

{{\cellcolor{mypink}{CLIP-Ap.}}} 
&\multicolumn{2}{|c}{{\cellcolor{mypink}${67.40}$\stdvno{$bs.$}}}
&\multicolumn{2}{c}{{\cellcolor{mypink}${48.10}$\stdvno{$bs.$}}}
&\multicolumn{2}{c}{{\cellcolor{mypink}$\underline{62.50}$\stdvno{$bs.$}}}
&\multicolumn{2}{c}{{\cellcolor{mypink}${72.90}$\stdvno{$bs.$}}}
&\multicolumn{2}{c}{{\cellcolor{mypink}${47.10}$\stdvno{$bs.$}}}
&\multicolumn{2}{|c}{{\cellcolor{mypink}${59.60}$\stdvno{$bs.$}}} 
&\multicolumn{2}{|c}{{\cellcolor{mypink}${57.65}$\stdvno{$bs.$}}} \\

{{\cellcolor{mypink}\textbf{$\text{IT}^{3}\text{A}$\&CLIP-Ap.}}} 
&\multicolumn{2}{|c}{{\cellcolor{mypink}$\textbf{68.60}$\stdvu{${{1.20}}$}}}
&\multicolumn{2}{c}{{\cellcolor{mypink}$\textbf{56.10}$\stdvu{${{8.00}}$}}}
&\multicolumn{2}{c}{{\cellcolor{mypink}$\textbf{63.20}$\stdvu{${{0.70}}$}}}
&\multicolumn{2}{c}{{\cellcolor{mypink}$\textbf{74.60}$\stdvu{${{1.70}}$}}}
&\multicolumn{2}{c}{{\cellcolor{mypink}$\textbf{50.30}$\stdvu{${{3.20}}$}}}
&\multicolumn{2}{|c}{{\cellcolor{mypink}$\textbf{62.56}$\stdvu{${{2.94}}$}}} 
&\multicolumn{2}{|c}{{\cellcolor{mypink}$\textbf{61.05}$\stdvu{${{3.40}}$}}} \\

\bottomrule
\end{tabular}
\label{tab:1}
\vspace{-10pt}
\end{table*}

\begin{figure*}[!t]
\renewcommand{\arraystretch}{1.3}
	\makeatletter\def\@captype{table}\makeatother\caption{{\textbf{Top 1 accuracy} $\%$ of state-of-the-art baselines under $\mathcal{S}_2$. \textbf{Avg.} represents the average performance of the \texttt{Cross-Dataset Generalization}. Arrows (${\color{ForestGreen}\uparrow}$ and ${\color{red}\downarrow}$) indicate enhancements and reductions compared to the CLIP method, \ie, CLIP-RN50 and CLIP-ViT-B/16. For comprehensive analyses, refer to Sec.~\ref{sec:acc}.}
}
	\label{tab:2}\centering
 \begin{minipage}[c]{\textwidth}
        
	\fontsize{8.6}{8.6}\selectfont
	\centering
	\begin{tabular}{l cc cc cc cc cc  }
\toprule
\centering
\textbf{Method}

&\multicolumn{2}{c}{\textbf{~~~~~~~Flower~~~~~~}}
&\multicolumn{2}{c}{\textbf{~~~~~~DTD~~~~~~}}
&\multicolumn{2}{c}{\textbf{~~~~~~Pets~~~~~~}} 
&\multicolumn{2}{c}{\textbf{~~~~~~Cars~~~~~~}}
&\multicolumn{2}{c}{\textbf{~~~~~~UCF101~~~~~~}} \\

\cmidrule(r){1-1}  \cmidrule(lr){2-3} \cmidrule(lr){4-5} \cmidrule(lr){6-7} \cmidrule(lr){8-9} \cmidrule(lr){10-11} 

{{\cellcolor{mygray}CLIP-RN50}} 
&\multicolumn{2}{|c}{{\cellcolor{mygray}{$61.60$\stdvno{$bs.$}}}}
&\multicolumn{2}{c}{{\cellcolor{mygray}{$38.50$\stdvno{$bs.$}} }}
&\multicolumn{2}{c}{{\cellcolor{mygray}{$84.70$\stdvno{$bs.$}} }}
&\multicolumn{2}{c}{{\cellcolor{mygray}{$\underline{55.70}$\stdvno{$bs.$}}}}
&\multicolumn{2}{c}{{\cellcolor{mygray}{$58.6$\stdvno{$bs.$}} }}\\

CoOp${\color{magenta}_{2022}}$~\citep{zhou2022learning}
&\multicolumn{2}{|c}{$60.90$}
&\multicolumn{2}{c}{$36.60$} 
&\multicolumn{2}{c}{$88.00$} 
&\multicolumn{2}{c}{$54.60$}
&\multicolumn{2}{c}{$59.01$}\\

CoCoOp${\color{magenta}_{2022}}$~\citep{zhou2022conditional}
&\multicolumn{2}{|c}{$\underline{63.90}$}
&\multicolumn{2}{c}{$\underline{40.70}$} 
&\multicolumn{2}{c}{$\underline{88.50}$} 
&\multicolumn{2}{c}{$53.50$}
&\multicolumn{2}{c}{$\underline{59.60}$} \\

CLIP-Ap.${\color{magenta}_{2024}}$~\citep{gao2024clip}
&\multicolumn{2}{|c}{$62.90$}
&\multicolumn{2}{c}{$40.10$} 
&\multicolumn{2}{c}{$84.80$} 
&\multicolumn{2}{c}{$55.10$}
&\multicolumn{2}{c}{$58.80$} \\

TPT${\color{magenta}_{2022}}$~\citep{shu2022test} 
&\multicolumn{2}{|c}{$59.20$\stdvd{${{2.40}}$}}
&\multicolumn{2}{c}{$39.10$\stdvu{${{0.60}}$}} 
&\multicolumn{2}{c}{$83.30$\stdvd{${{1.40}}$}} 
&\multicolumn{2}{c}{$54.90$\stdvd{${{0.80}}$}}
&\multicolumn{2}{c}{$59.6$\stdvu{${{1.00}}$}} \\

{{\cellcolor{mypink}{DiffTPT}${\color{magenta}_{2023}}$}~\citep{feng2023diverse}} 
&\multicolumn{2}{|c}{{\cellcolor{mypink}${57.10}$\stdvd{${{4.50}}$}}}
&\multicolumn{2}{c}{{\cellcolor{mypink}${38.60}$\stdvu{${{0.10}}$}}}
&\multicolumn{2}{c}{{\cellcolor{mypink}${84.20}$\stdvd{${{0.50}}$}}}
&\multicolumn{2}{c}{{\cellcolor{mypink}$\textbf{57.30}$\stdvu{${{1.60}}$}}}
&\multicolumn{2}{c}{{\cellcolor{mypink}$\textbf{63.70}$\stdvu{${{5.10}}$}}}\\

{{\cellcolor{mypink}\textbf{$\text{IT}^{3}\text{A}$}}} 
&\multicolumn{2}{|c}{{\cellcolor{mypink}$\textbf{59.40}$\stdvd{${{1.20}}$}}}
&\multicolumn{2}{c}{{\cellcolor{mypink}$\textbf{42.30}$\stdvu{${{3.80}}$}}}
&\multicolumn{2}{c}{{\cellcolor{mypink}$\textbf{85.40}$\stdvu{${{0.70}}$}}}
&\multicolumn{2}{c}{{\cellcolor{mypink}$\textbf{57.30}$\stdvu{${{1.60}}$}}}
&\multicolumn{2}{c}{{\cellcolor{mypink}${63.40}$\stdvu{${{4.80}}$}}}\\


\bottomrule
\end{tabular}
\end{minipage}


\centering
 \begin{minipage}[c]{\textwidth}\centering
        \centering
	\fontsize{8.6}{8.6}\selectfont
	\centering
	\begin{tabular}{l cc cc cc cc cc cc  }
\toprule
\centering
\textbf{Method}

&\multicolumn{2}{c}{\textbf{~Caltech11~}} 
&\multicolumn{2}{c}{\textbf{~Food101~}}
&\multicolumn{2}{c}{\textbf{SUN397~}}
&\multicolumn{2}{c}{\textbf{~Aircraft~}}
&\multicolumn{2}{c}{\textbf{~EuroSAT~}}
&\multicolumn{2}{c}{\textbf{Avg.}} \\

\cmidrule(r){1-1} 
\cmidrule(lr){2-3} \cmidrule(lr){4-5} \cmidrule(lr){6-7} \cmidrule(lr){8-9} \cmidrule(lr){10-11} \cmidrule(lr){12-13} 

{{\cellcolor{mygray}CLIP-RN50}} 
&\multicolumn{2}{|c}{{\cellcolor{mygray}{$85.20$\stdvno{$bs.$}} }}
&\multicolumn{2}{c}{{\cellcolor{mygray}{$75.90$\stdvno{$bs.$}}}} 
&\multicolumn{2}{c}{{\cellcolor{mygray}{$60.00$\stdvno{$bs.$}}}}
&\multicolumn{2}{c}{{\cellcolor{mygray}{$15.50$\stdvno{$bs.$}} }}
&\multicolumn{2}{c}{{\cellcolor{mygray}{$19.70$\stdvno{$bs.$}} }}
&\multicolumn{2}{|c}{{\cellcolor{mygray}{$55.54$\stdvno{$bs.$}}}} 

\\

CoOp${\color{magenta}_{2022}}$~\citep{zhou2022learning}
&\multicolumn{2}{|c}{$86.10$} 
&\multicolumn{2}{c}{$78.20$}
&\multicolumn{2}{c}{$59.00$}
&\multicolumn{2}{c}{$16.10$} 
&\multicolumn{2}{c}{$22.80$} 
&\multicolumn{2}{|c}{$56.24$} \\

CoCoOp${\color{magenta}_{2022}}$~\citep{zhou2022conditional}
&\multicolumn{2}{|c}{$\underline{87.70}$} 
&\multicolumn{2}{c}{$\underline{78.50}$}
&\multicolumn{2}{c}{$59.60$}
&\multicolumn{2}{c}{$15.40$} 
&\multicolumn{2}{c}{$\underline{30.50}$} 
&\multicolumn{2}{|c}{$\underline{57.79}$} \\

CLIP-Ap.${\color{magenta}_{2024}}$~\citep{gao2024clip}
&\multicolumn{2}{|c}{$86.10$} 
&\multicolumn{2}{c}{$74.20$}
&\multicolumn{2}{c}{$\underline{60.70}$}
&\multicolumn{2}{c}{$\underline{16.60}$} 
&\multicolumn{2}{c}{$25.80$} 
&\multicolumn{2}{|c}{$56.51$} \\

TPT${\color{magenta}_{2022}}$~\citep{shu2022test}
&\multicolumn{2}{|c}{$84.30$\stdvd{${{0.90}}$}} 
&\multicolumn{2}{c}{$75.80$\stdvd{${{0.10}}$}}
&\multicolumn{2}{c}{$60.90$\stdvu{${{0.90}}$}}
&\multicolumn{2}{c}{$17.00$\stdvu{${{1.50}}$}} 
&\multicolumn{2}{c}{$22.80$\stdvu{${{3.10}}$}} 
&\multicolumn{2}{|c}{$55.69$\stdvu{${{0.15}}$}} \\

{{\cellcolor{mypink}{DiffTPT}${\color{magenta}_{2023}}$}~\citep{feng2023diverse}} 
&\multicolumn{2}{|c}{{\cellcolor{mypink}${87.30}$\stdvu{${{2.10}}$}}} 
&\multicolumn{2}{c}{{\cellcolor{mypink}$\textbf{75.90}$\stdvu{${{0.00}}$}}}
&\multicolumn{2}{c}{{\cellcolor{mypink}$\textbf{63.10}$\stdvu{${{3.10}}$}}}
&\multicolumn{2}{c}{{\cellcolor{mypink}${16.50}$\stdvu{${{1.00}}$}}}
&\multicolumn{2}{c}{{\cellcolor{mypink}${34.50}$\stdvu{${{14.80}}$}}} 
&\multicolumn{2}{|c}{{\cellcolor{mypink}${57.82}$\stdvu{${{2.28}}$}}} \\

{{\cellcolor{mypink}\textbf{$\text{IT}^{3}\text{A}$}}} 
&\multicolumn{2}{|c}{{\cellcolor{mypink}$\textbf{87.60}$\stdvu{${{2.40}}$}}} 
&\multicolumn{2}{c}{{\cellcolor{mypink}${74.20}$\stdvd{${{1.70}}$}}}
&\multicolumn{2}{c}{{\cellcolor{mypink}${60.90}$\stdvu{${{0.90}}$}}}
&\multicolumn{2}{c}{{\cellcolor{mypink}$\textbf{18.30}$\stdvu{${{2.50}}$}}}
&\multicolumn{2}{c}{{\cellcolor{mypink}$\textbf{40.00}$\stdvu{${{21.30}}$}}} 
&\multicolumn{2}{|c}{{\cellcolor{mypink}$\textbf{58.88}$\stdvu{${{3.34}}$}}} \\

\bottomrule
\end{tabular}
\end{minipage} 

 \begin{minipage}[c]{\textwidth}
        
	\fontsize{8.6}{8.6}\selectfont
	\centering
	\begin{tabular}{l cc cc cc cc cc  }
\toprule
\centering
\textbf{Method}

&\multicolumn{2}{c}{\textbf{~~~~~~~Flower~~~~~~}}
&\multicolumn{2}{c}{\textbf{~~~~~~DTD~~~~~~}}
&\multicolumn{2}{c}{\textbf{~~~~~~Pets~~~~~~}} 
&\multicolumn{2}{c}{\textbf{~~~~~~Cars~~~~~~}}
&\multicolumn{2}{c}{\textbf{~~~~~~UCF101~~~~~~}} \\

\cmidrule(r){1-1}  \cmidrule(lr){2-3} \cmidrule(lr){4-5} \cmidrule(lr){6-7} \cmidrule(lr){8-9} \cmidrule(lr){10-11}

{{\cellcolor{mygray}CLIP-ViT-B/16}} 
&\multicolumn{2}{|c}{{\cellcolor{mygray}{$66.50$\stdvno{$bs.$} }}}
&\multicolumn{2}{c}{{\cellcolor{mygray}{$41.90$\stdvno{$bs.$} }}} 
&\multicolumn{2}{c}{{\cellcolor{mygray}{$88.60$\stdvno{$bs.$}}}} 
&\multicolumn{2}{c}{{\cellcolor{mygray}{$\underline{66.80}$\stdvno{$bs.$}}}}
&\multicolumn{2}{c}{{\cellcolor{mygray}{$63.70$\stdvno{$bs.$}} }}\\

CoOp${\color{magenta}_{2022}}$~\citep{zhou2022learning} 
&\multicolumn{2}{|c}{$\underline{68.10}$}
&\multicolumn{2}{c}{$41.60$} 
&\multicolumn{2}{c}{$89.80$} 
&\multicolumn{2}{c}{$65.30$}
&\multicolumn{2}{c}{$64.50$} \\

CoCoOp${\color{magenta}_{2022}}$~\citep{zhou2022conditional} 
&\multicolumn{2}{|c}{$65.70$}
&\multicolumn{2}{c}{$42.00$} 
&\multicolumn{2}{c}{$\underline{90.00}$} 
&\multicolumn{2}{c}{$60.80$}
&\multicolumn{2}{c}{$61.20$}  \\

CLIP-Ap.${\color{magenta}_{2024}}$~\citep{gao2024clip}
&\multicolumn{2}{|c}{$67.40$}
&\multicolumn{2}{c}{$\underline{43.20}$} 
&\multicolumn{2}{c}{$89.20$} 
&\multicolumn{2}{c}{$65.90$}
&\multicolumn{2}{c}{$\underline{64.60}$}  \\

TPT${\color{magenta}_{2022}}$~\citep{shu2022test} 
&\multicolumn{2}{|c}{$66.50$\stdvu{${{0}}$}}
&\multicolumn{2}{c}{$43.10$\stdvu{${{1.20}}$}} 
&\multicolumn{2}{c}{$86.80$\stdvd{${{1.80}}$}} 
&\multicolumn{2}{c}{$66.50$\stdvd{${{0.30}}$}}
&\multicolumn{2}{c}{$67.80$\stdvu{${{4.10}}$}} \\

{{\cellcolor{mypink}{DiffTPT}${\color{magenta}_{2023}}$}~\citep{feng2023diverse}} 
&\multicolumn{2}{|c}{{\cellcolor{mypink}${67.20}$\stdvu{${{0.70}}$}}}
&\multicolumn{2}{c}{{\cellcolor{mypink}${43.50}$\stdvu{${{1.60}}$}}}
&\multicolumn{2}{c}{{\cellcolor{mypink}${85.90}$\stdvd{${{2.70}}$}}}
&\multicolumn{2}{c}{{\cellcolor{mypink}${65.90}$\stdvd{${{0.90}}$}}}
&\multicolumn{2}{c}{{\cellcolor{mypink}${66.50}$\stdvu{${{2.80}}$}}}\\

{{\cellcolor{mypink}\textbf{$\text{IT}^{3}\text{A}$}}} 
&\multicolumn{2}{|c}{{\cellcolor{mypink}$\textbf{69.90}$\stdvu{${{3.40}}$}}}
&\multicolumn{2}{c}{{\cellcolor{mypink}$\textbf{44.50}$\stdvu{${{2.60}}$}}}
&\multicolumn{2}{c}{{\cellcolor{mypink}$\textbf{88.80}$\stdvu{${{0.20}}$}}}
&\multicolumn{2}{c}{{\cellcolor{mypink}$\textbf{66.90}$\stdvu{${{0.10}}$}}}
&\multicolumn{2}{c}{{\cellcolor{mypink}$\textbf{69.00}$\stdvu{${{5.30}}$}}}\\

\bottomrule
\end{tabular}
\end{minipage}


\centering
 \begin{minipage}[c]{\textwidth}\centering
        \centering
	\fontsize{8.6}{8.6}\selectfont
	\centering
	\begin{tabular}{l cc cc cc cc cc cc  }
\toprule
\centering
\textbf{Method}

&\multicolumn{2}{c}{\textbf{~Caltech11~}} 
&\multicolumn{2}{c}{\textbf{~Food101~}}
&\multicolumn{2}{c}{\textbf{SUN397~}}
&\multicolumn{2}{c}{\textbf{~Aircraft~}}
&\multicolumn{2}{c}{\textbf{~EuroSAT~}}
&\multicolumn{2}{c}{\textbf{Avg.}} \\

\cmidrule(r){1-1} 
\cmidrule(lr){2-3} \cmidrule(lr){4-5} \cmidrule(lr){6-7} \cmidrule(lr){8-9} \cmidrule(lr){10-11} \cmidrule(lr){12-13} 

{{\cellcolor{mygray}CLIP-ViT-B/16}} 
&\multicolumn{2}{|c}{{\cellcolor{mygray}{$91.90$\stdvno{$bs.$}}}} 
&\multicolumn{2}{c}{{\cellcolor{mygray}{$85.40$\stdvno{$bs.$}}}}
&\multicolumn{2}{c}{{\cellcolor{mygray}{$64.10$\stdvno{$bs.$}}}}
&\multicolumn{2}{c}{{\cellcolor{mygray}{$\underline{24.00}$\stdvno{$bs.$}} }}
&\multicolumn{2}{c}{{\cellcolor{mygray}{$40.60$\stdvno{$bs.$}}}} 
&\multicolumn{2}{|c}{{\cellcolor{mygray}{$63.35$\stdvno{$bs.$}}}}\\

CoOp${\color{magenta}_{2022}}$~\citep{zhou2022learning}
&\multicolumn{2}{|c}{$91.80$} 
&\multicolumn{2}{c}{$83.80$}
&\multicolumn{2}{c}{$64.60$}
&\multicolumn{2}{c}{$17.60$} 
&\multicolumn{2}{c}{$32.00$} 
&\multicolumn{2}{|c}{$61.91$} \\

CoCoOp${\color{magenta}_{2022}}$~\citep{zhou2022conditional}
&\multicolumn{2}{|c}{$90.80$} 
&\multicolumn{2}{c}{$\underline{85.50}$}
&\multicolumn{2}{c}{$64.00$}
&\multicolumn{2}{c}{$16.00$} 
&\multicolumn{2}{c}{$\underline{44.80}$} 
&\multicolumn{2}{|c}{$62.08$} \\

CLIP-Ap.${\color{magenta}_{2024}}$~\citep{gao2024clip}
&\multicolumn{2}{|c}{$\underline{92.30}$} 
&\multicolumn{2}{c}{$84.80$}
&\multicolumn{2}{c}{$\underline{65.30}$}
&\multicolumn{2}{c}{$23.10$} 
&\multicolumn{2}{c}{$39.40$} 
&\multicolumn{2}{|c}{$\underline{63.50}$} \\

TPT${\color{magenta}_{2022}}$~\citep{shu2022test}
&\multicolumn{2}{|c}{$91.50$\stdvd{${{0.40}}$}} 
&\multicolumn{2}{c}{$\textbf{86.20}$\stdvu{${{0.80}}$}}
&\multicolumn{2}{c}{$66.20$\stdvu{${{2.10}}$}}
&\multicolumn{2}{c}{$21.20$\stdvd{${{2.80}}$}} 
&\multicolumn{2}{c}{$37.00$\stdvd{${{3.60}}$}} 
&\multicolumn{2}{|c}{$63.28$\stdvd{${{0.07}}$}} \\

{{\cellcolor{mypink}{DiffTPT}${\color{magenta}_{2023}}$}~\citep{feng2023diverse}}  
&\multicolumn{2}{|c}{{\cellcolor{mypink}$\textbf{94.00}$\stdvu{${{2.10}}$}}} 
&\multicolumn{2}{c}{{\cellcolor{mypink}${84.40}$\stdvd{${{1.00}}$}}}
&\multicolumn{2}{c}{{\cellcolor{mypink}${67.30}$\stdvu{${{3.20}}$}}}
&\multicolumn{2}{c}{{\cellcolor{mypink}${20.50}$\stdvd{${{3.50}}$}}}
&\multicolumn{2}{c}{{\cellcolor{mypink}$\textbf{41.60}$\stdvu{${{1.00}}$}}} 
&\multicolumn{2}{|c}{{\cellcolor{mypink}${63.68}$\stdvu{${{0.33}}$}}} \\

{{\cellcolor{mypink}\textbf{$\text{IT}^{3}\text{A}$}}} 
&\multicolumn{2}{|c}{{\cellcolor{mypink}${93.80}$\stdvu{${{1.90}}$}}} 
&\multicolumn{2}{c}{{\cellcolor{mypink}${84.50}$\stdvd{${{0.90}}$}}}
&\multicolumn{2}{c}{{\cellcolor{mypink}$\textbf{68.80}$\stdvu{${{4.70}}$}}}
&\multicolumn{2}{c}{{\cellcolor{mypink}$\textbf{25.40}$\stdvu{${{1.40}}$}}}
&\multicolumn{2}{c}{{\cellcolor{mypink}${39.70}$\stdvd{${{0.90}}$}}} 
&\multicolumn{2}{|c}{{\cellcolor{mypink}$\textbf{65.13}$\stdvu{${{1.78}}$}}} \\

\bottomrule
\end{tabular}
\end{minipage} 

\end{figure*}

\subsection{Comparison with State-of-the-arts}\label{sec:acc}

\subsubsection{{Natural Distribution Shifts}}
{
Table~\ref{tab:1} provides an overview of the performance assessments for various competitive approaches within $\mathcal{S}$cenario 1, utilizing different backbones such as ResNet-50 and ViT-B/16. In this context, CLIP represents the zero-shot CLIP output with the standard prompt ``\texttt{a photo of a}''.
``---\&CoOp'' and ``---\&CoCoOp'' refer to the implementation of test-time prompt tuning techniques on CoOp~\citep{zhou2022learning} or CoCoOp~\citep{zhou2022conditional}, respectively. These methods are fine-tuned with $16$-shot training samples per category on ImageNet.
Our proposed $\text{IT}^{3}\text{A}$ along with CLIP-Adapter and $\text{IT}^{3}\text{A}$ initialized with few-shot pretrained CLIP-Adapter weights is also included in the table. We note that our method aims to maintain effective performance with limited augmented images in a resource-efficient manner. Therefore, unlike the conference version of DiffTPT~\citep{feng2023diverse}, we only use $8$-fold augmentation here.
As demonstrated in the table, in general $\text{IT}^{3}\text{A}$ and $\text{IT}^{3}\text{A\&CLIP-Ap.}$ outperform all other methods and their variants correspondingly.
On CLIP-RN50, the average performance of $\text{IT}^{3}\text{A}$ improved by $5.50$, and the average for OOD generation was improved by $6.22$. Compared to the conference version of DiffTPT, the performances are as follows: DiffTPT: $46.70$/$43.88$ \textit{vs.} $\text{IT}^{3}\text{A}$: \textbf{49.26}/\textbf{46.75}.
On CLIP-Adapter, $\text{IT}^{3}\text{A}$ also showed significant improvements, with all its performances exceeding those of DiffTPT applied to CoOp~\citep{zhou2022learning} or CoCoOp~\citep{zhou2022conditional}. Both methods enhance in-domain accuracy on \textbf{ImageNet} data and generalization to OOD data. For ResNet-50-based in-domain average performance, DiffTPT\&CoCoOp: $45.94$ \textit{vs.} $\text{IT}^{3}\text{A}$\&CLIP-Adapter: \textbf{49.94}; for the generalization test of OOD data, DiffTPT\&CoCoOp: $42.88$ \textit{vs.} $\text{IT}^{3}\text{A}$\&CLIP-Adapter: \textbf{46.83}.
Our method also achieved similar improvements on ViT-B/16. Specifically, compared to DiffTPT: $60.24$ $\rightarrow$ $\textbf{60.60}$, $59.10$ $\rightarrow$ $\textbf{59.25}$, $60.20$ $\rightarrow$ $\textbf{62.56}$, and $58.53$ $\rightarrow$ $\textbf{61.05}$.
Since TPT~\citep{shu2022test} use random resized cropping to augment test images, their generalization ability is limited. DiffTPT~\citep{feng2023diverse} can only acquire visual diversity from a single image modality, limiting the knowledge it captures.
Notably, we found that $\text{IT}^{3}\text{A}$ significantly improves the generalization test on OOD data. This supports our conclusion that $\text{IT}^{3}\text{A}$ enhances robustness by acquiring more knowledge through multi-modal augmentation (\ie, \textit{\textbf{prediction fidelity}}) and increasing the \textit{\textbf{data diversity}} of the test samples.
}

\begin{figure*}[t]
	\begin{center}
		\includegraphics[width=0.98\linewidth]{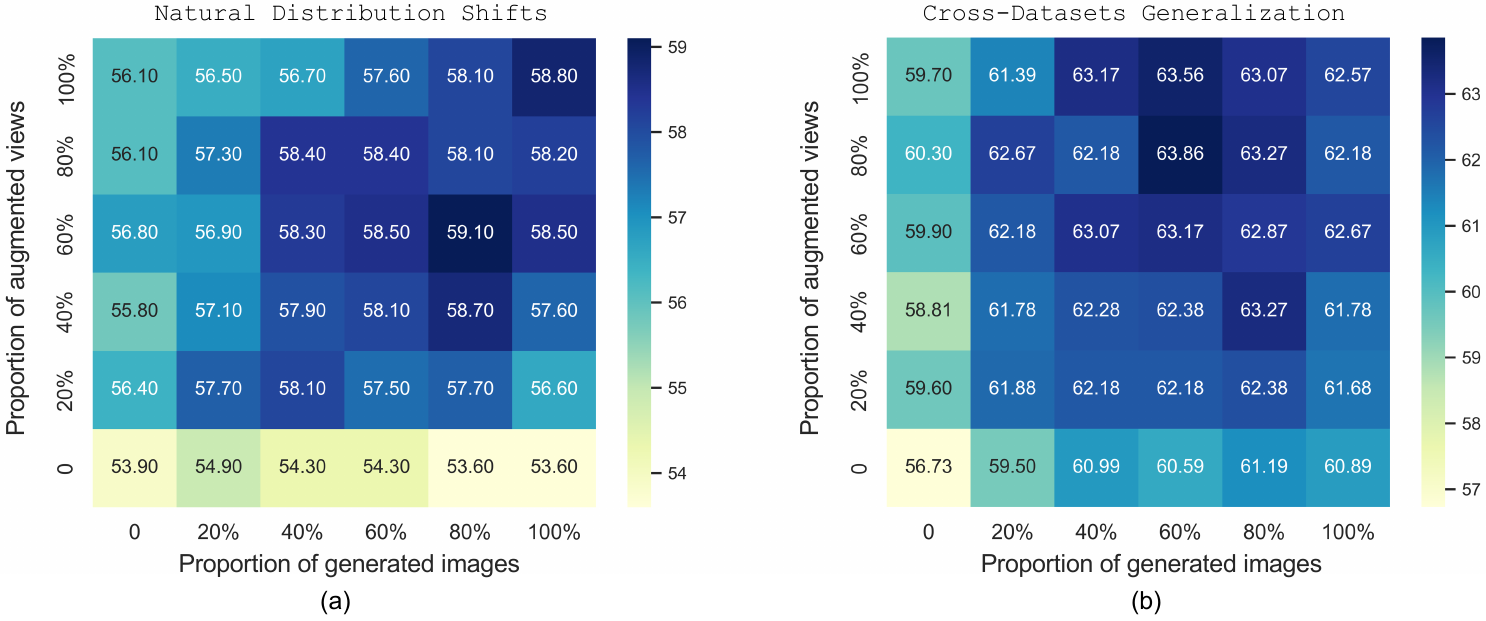}
	\end{center}
	\captionsetup{font=small}
    	\caption{{\textbf{Top-1 accuracy variation} against different proportions of standard augmented views and diffusion-augmented images for scenarios (a) $\mathcal{S}_1$ and (b) $\mathcal{S}_2$.}}   
	\label{fig:3}
\end{figure*}

\begin{figure*}[t]
	\begin{center}
		\includegraphics[width=\linewidth]{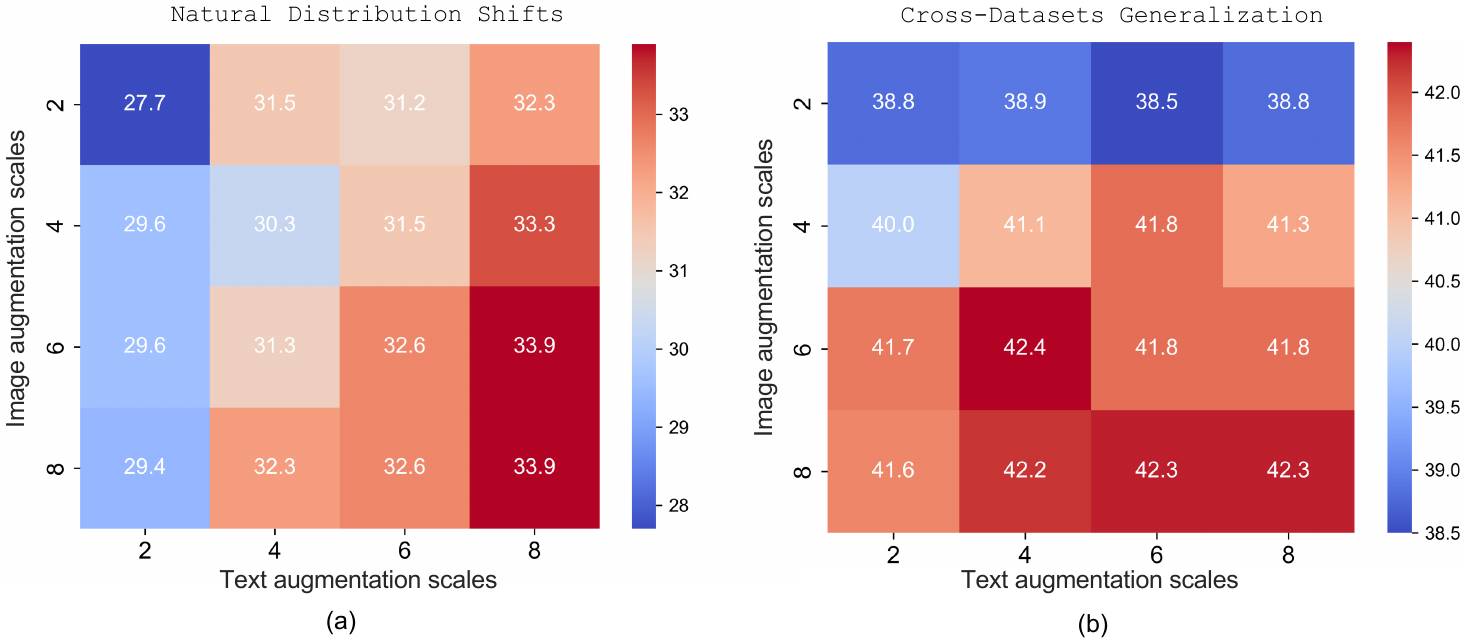}
	\end{center}
	\captionsetup{font=small}
    	\caption{{\textbf{Top-1 accuracy variation} against different scales of text augmentation and image augmentation for (a) ImageNet-A under scenario $\mathcal{S}_1$ and (b) DTD under scenario $\mathcal{S}_2$.}} 
	\label{fig:re2}
\end{figure*}


{
Naturally, the results generated by CLIP are the lowest, as direct testing on new datasets is significantly impacted by domain shifts. 
Although CoOp~\citep{zhou2022learning} and CoCoOp~\citep{zhou2022conditional} benefit from learnable prompts, these methods depend on training datasets and do not utilize prompt tuning at test time. CLIP-Adapter achieves performance gain from learnable adapters but also faces the same problem. This means said methods fail to consider zero-shot generalization in practical, real-world settings, which results in lower effective performance.
Our findings confirm the initial hypothesis that enhancing test data with diverse synthetic data can boost zero-shot generalization performance.
}

\subsubsection{{Cross-Dataset Generalization.}}
We evaluate the ability of our proposed method and several baseline models to generalize from ImageNet to $10$ different fine-grained datasets by recording their quantitative performances, as shown in Table~\ref{tab:2}. TPT~\citep{shu2022test} is deployed in a zero-shot manner, CoOp~\citep{zhou2022learning}, CoCoOp~\citep{zhou2022conditional}, and CLIP-Adapter~\citep{gao2024clip} are fine-tuned on \textbf{ImageNet} with $16$-shot samples per category. Due to the diverse nature of these fine-grained datasets, each method displays varying performance levels on each dataset.
However, our proposed $\text{IT}^{3}\text{A}$ still achieves the best performance, \ie, raising the \textbf{Avg.} accuracy from $55.54$ to $\textbf{58.88}$ on CLIP-RN50, and from $63.03$ to $\textbf{65.13}$ on CLIP-ViT-B/16, all based on just 8-fold multi-modal augmentation.
Notably, our method achieved performance gains of $1.06\%$ and $1.45\%$ over DiffTPT \citep{shu2022test} on ResNet-50 and ViT-B/16 backbones, respectively.
This demonstrates that, among all competing methods, our approach is robust to natural distribution shifts even without training data, and significantly outperforms few-shot prompt tuning methods, \ie, CoOp \citep{zhou2022learning}, CoCoOp \citep{zhou2022conditional}, and CLIP-Adapter~\citep{gao2024clip}.
Although TPT, DiffTPT, and our proposed $\text{IT}^{3}\text{A}$ exhibits some performance decline on a few datasets, this is primarily due to the limited augmentation used during the testing phase, \ie, 8-fold, while in the previous version of TPT and DiffTPT, they augment the test with 64-fold.

\subsection{Ablation Studies}\label{sec:ab}
\subsubsection{{Balancing Synthetic Data \textit{vs.}~Standard Augmentation}}~
Given that our method for image domain augmentation leverages the complementary advantages of both the standard augmentation~\citep{shu2022test} and diffusion-based augmentation, it is essential to investigate how these two methods contribute to training the classifier.
To address this, we assess the average performance of ResNet-50 across the two scenarios, \ie, $\mathcal{S}_1$ on ImageNet-R and $\mathcal{S}_2$ on UCF101, inherit the conference version.
For improved visualization, we present a plot in Fig.~\ref{fig:3} that illustrates mixed combinations of different ratios, with the x-axis representing the percentage of synthetic data generated through diffusion-based augmentation and the y-axis indicating the percentage of data obtained through standard augmentation.
In the matrix of this figure, each cell $\mathcal{V}_{ij}$ corresponds to the classification performance of DiffTPT using $i$\% of synthetic data and $j$\% of standard augmented data.
Fig.~\ref{fig:3} (a) demonstrates a significant improvement in accuracy for \texttt{Natural Distribution Shifts} as the quantity of standard augmented data increases while keeping the synthetic data level fixed. In contrast, the effects are even more pronounced when the proportion of synthetic data is increased while the amount of standard augmented data is held constant. Overall, increasing the amount of synthetic data leads to better performance in $\mathcal{S}_1$. 
In Fig.~\ref{fig:3} (b), we show the performance of the classifier for $\mathcal{S}_2$, \ie, \texttt{Cross-Dataset Generalization}. We observe that, while keeping the amount of synthetic data fixed, the effectiveness of the classifier increases significantly as the proportion of standard augmented data increases.

\begin{figure}[t]
	\begin{center}
		\includegraphics[width=\linewidth]{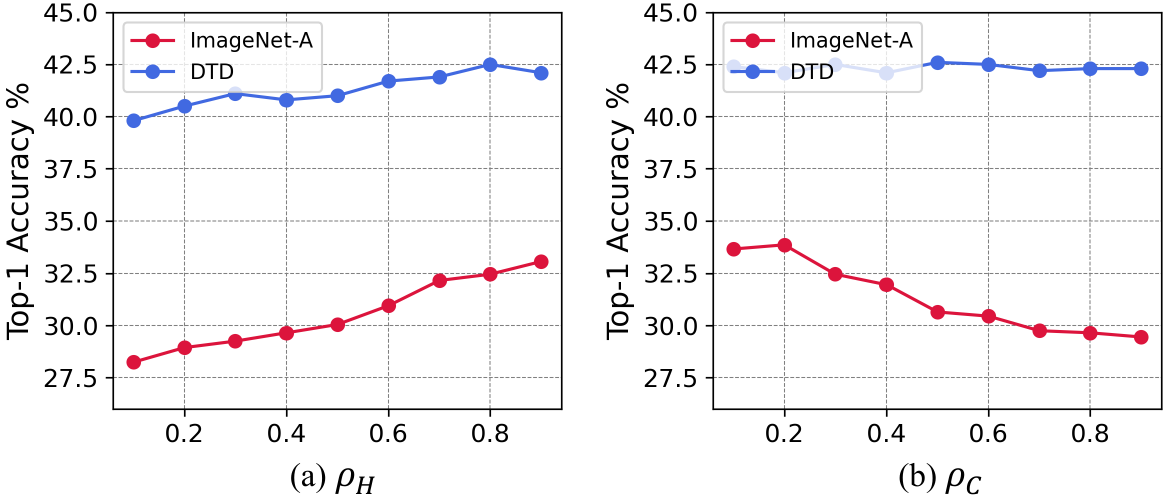}
	\end{center}
	\captionsetup{font=small}
	\caption{\small{Top-1 accuracy analysis of the \textbf{ratios $\rho_H$} and $\rho_C$ with regard to $\mathcal{S}_1$ and $\mathcal{S}_2$.}}
	\label{fig:4}
\end{figure}

\subsubsection{{Analysis of Ratio $\rho_H$ and $\rho_C$}}
As mentioned in Sec.~\ref{sec:cosine}, $\rho_H$ and $\rho_C$ filter out the less informative ``noisy'' and spurious augmented pairs in overall generative augmentation by standard of self-entropy and cosine similarity.
We evaluated the accuracy for various values of $\rho_H$ and $\rho_C$ in Fig.~\ref{fig:4} across two scenarios to determine the amount of information that good test augmentations should retain. From Fig.~\ref{fig:4} (a), it can be observed that on the DTD dataset under $\mathcal{S}_2$, there is a trade-off between augmented data quantity and augmented data quality, with the highest accuracy achieved at a value of $0.8$. 
%
Compared to $\mathcal{S}_2$, $\mathcal{S}_1$ requires more data for extended amount of learning to bridge the gap between in-domain ImageNet distribution and OOD adversarial samples from ImageNet-A. Accordingly, the performance of $\text{IT}^{3}\text{A}$ improves with an increase in $\rho_H$ on the ImageNet-A dataset.
For $\rho_C$ in Fig.~\ref{fig:4} (b), larger values correspond to more data pairs, while smaller values indicate higher data quality. The Fig.~\ref{fig:4} (b) illustrates a trade-off between the number of augmented data pairs and data quality.



\subsubsection{{Effect of the Generated Dataset Size}}
Since the primary contribution of our method lies in integrating multi-modal augmentation information, we investigate the impact of these two modalities, \ie, image and text, on classifier training. In Fig.~\ref{fig:re2}, we present mixed combinations of different modalities and augmentation scales, where the x-axis represents the scale of text augmentation and the y-axis represents the scale of image augmentation.
In the matrix of this figure, each element $\mathcal{V}_{ij}$ indicates the classification performance of $\text{IT}^{3}\text{A}$ with $i \times$ text augmentation and $j \times$ image augmentation.
As illustrated in Fig.~\ref{fig:re2} (a), in $\mathcal{S}_1$, the augmented textual information provides essential components for domain adaptation. Consequently, as the augmentation levels for both text and images increase, the model performance improves. In contrast, for $\mathcal{S}_2$, the model relies less on textual data, resulting in reduced sensitivity to changes in text augmentation levels.

\begin{figure}[t]
	\begin{center}
		\includegraphics[width=0.8\linewidth]{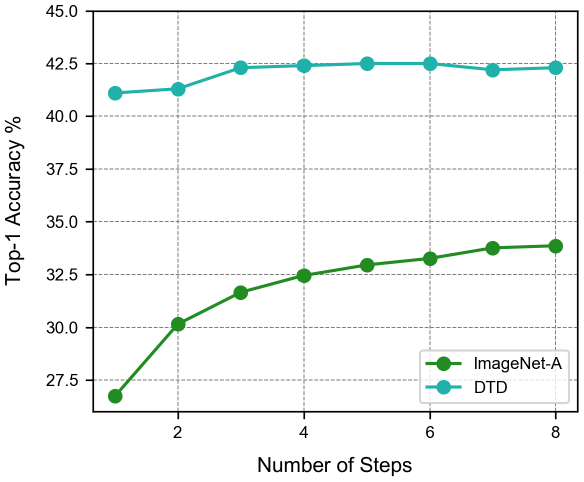}
	\end{center}
	\captionsetup{font=small}
    	\caption{{\textbf{Ablation studies} on the \textit{learning steps} under $\mathcal{S}_1$ and $\mathcal{S}_2$.}}
	\label{fig:6}
\end{figure}

\subsubsection{{Steps of Prompt Updating}}
To evaluate the effectiveness of the learning updates, we have recorded the accuracy across different optimization steps for two scenarios in Fig.~\ref{fig:6}.
As illustrated in the figure, the performance of $\text{IT}^{3}\text{A}$ on the DTD dataset under $\mathcal{S}_2$ continually improves from $27.3$ to $\textbf{33.8}$ with an increasing number of optimization steps.
In contrast, on the ImageNet-A dataset under $\mathcal{S}_1$, the performance of $\text{IT}^{3}\text{A}$ increases to $42.5$ and then stabilizes as the number of optimization steps increases.
This indicates that additional optimization steps do not provide further benefits to the classifier and only serve to increase inference time.
Taking both performance and computational efficiency into account, we set the number of steps to $4$ in our experiments.

\subsubsection{Effect of the Configurations of the Adapter}
Here, we investigate whether the different configurations will influence the model performance. Table~\ref{tab:layer} summarized the accuracies with different layers in the adapter on two scenarios, \ie, ImageNet-A under $\mathcal{S}_1$ and DTD under $\mathcal{S}_2$. As can be seen in the table, the performance for ImageNet-A dataset increases as the layers of adapter module increase from 1 layer to 2 layers, \ie, from $32.60$ to $33.90$, but then decreases to $32.00$ eventually as layers keep increasing to 8. Similarly, there is also a peak in performance in the middle of the layer range, \ie, $42.70$ at 4 layers. According to the results of different configurations of the adapter in this table, we set the layers to 2 in our method.

\begin{table}[t]
\centering
		\caption{Investigation of the different configurations of the adapter under both $\mathcal{S}_1$ and $\mathcal{S}_2$.}
        \fontsize{9}{12}\selectfont
        \setlength{\tabcolsep}{8pt}
			\begin{tabular}{l|cccc}
				\toprule
				\textbf{Datasets}
                &{1 layer}
                &{2 layers}
                &{4 layers} 
                &{8 layers} 
                \\ 
 
\cmidrule(lr){1-1}  
\cmidrule(lr){2-2} 
\cmidrule(lr){3-3} 
\cmidrule(lr){4-4} 
\cmidrule(lr){5-5} 

\textbf{ImageNet-A}
&$32.60$
&$33.90$
&$32.90$ 
&$32.00$ 
\\
     
\textbf{DTD}
&$42.00$
&$42.30$
&$42.70$ 
&$42.00$ 
\\

\bottomrule
\end{tabular}
\label{tab:layer}

\end{table}

\subsubsection{{Inference cost}}
Despite the fact that the original SD can be time-intensive, such as taking $6$ seconds to infer $10$ test images for TPT and $36$ minutes with standard SD, recent advancements have led to the development of faster SD models.
For example, ToMe~\citep{bolya2023token}, two-stage distillation~\citep{meng2023distillation}, and Consistency Model~\citep{song2023consistency}. Notably, the Consistency Model can generate $10$ images in just $0.5$ seconds, compared to the original SD’s $70$ seconds. 
Moreover, efficiency can be further enhanced using techniques like TensorRT and Memory Efficient Attention\footnote{https://www.photoroom.com/tech/stable-diffusion-100-percent-faster-with-memory-efficient-attention}, resulting in additional gains of $25$\% and $100$\% in inference speed, respectively.
Compared to DiffTPT, $\text{IT}^{3}\text{A}$ only needs to generate $\frac{1}{M}$ the number of images compared to DiffTPT for the same number of image-text pairs. Also, the computational cost of generating multiple distinct text templates through instructions for GPT-4 is negligible. Therefore, for approximately the same amount of computational consumption, $\text{IT}^{3}\text{A}$ can generate $M$ times of augment pairs of DiffTPT. In other words, the overall computational cost of $\text{IT}^{3}\text{A}$ is much lower than that of DiffTPT when comparing under the same amount of augment pairs.
Additionally, compared to the prompt-learning method used in the conference version, DiffTPT~\citep{feng2023diverse}, $\text{IT}^{3}\text{A}$ requires less tuning time for a single test image, \ie, $\text{IT}^{3}\text{A}$: $0.33$s \textit{vs}. DiffTPT: $1.08$s.

\section{Conclusion}
This paper proposes a multi-modal test-time optimization method that leverages enhanced data from pre-trained models in both image and text modalities. By combining the strengths of these modalities, the method improves the model's adaptability to unknown test data. To fully utilize the diversity provided by generative models in both vision and language, we have replaced prompt tuning from the conference version, DiffTPT, with adapters. This change allows for more flexible use of text templates on the text encoder. Additionally, using cosine similarity filtering between the original test data and the augmented images and text ensures that key semantics are faithfully preserved during various visual and textual augmentations. Experiments on test datasets with distribution shifts and unseen classes demonstrate that the $\text{IT}^{3}\text{A}$ method improves zero-shot accuracy by an average of 4.98\% compared to the state-of-the-art TPT method.
Our approach of multi-modal augmentation during test time can inspire developments in test time strategies for other multi-modal tasks, such as composed image retrieval, which is also a future direction we are currently exploring.

\noindent\textbf{Data Availability Statements.} The authors declare that the data supporting the experiments in this study are available within the paper. The code is available at \texttt{https://github.com/chunmeifeng/DiffTPT}.

\begin{acknowledgements}
This work was supported by the National Research Foundation, Singapore under its AI Singapore Programme (AISG Award No: AISG2-TC-2021-003), Agency for Science, Technology and Research (A*STAR) through its AME Programmatic Funding Scheme Under Project A20H4b0141, A*STAR Central Research Fund ``A Secure and Privacy Preserving AI Platform for Digital Health”, and Agency for Science, Technology and Research (A*STAR) through its RIE2020 Health and Biomedical Sciences (HBMS) Industry Alignment Fund Pre-Positioning (IAF-PP) (grant no. H20C6a0032), and Shenzhen-Hong Kong Joint Funding No. SGDX20211123112401002, by Shenzhen General Program No. JCYJ20220530143600001.

\end{acknowledgements}

\bibliographystyle{spbasic}      
\bibliography{sn-bibliography.bib}   

\begin{thebibliography}{77}
\providecommand{\natexlab}[1]{#1}
\providecommand{\url}[1]{{#1}}
\providecommand{\urlprefix}{URL }
\expandafter\ifx\csname urlstyle\endcsname\relax
  \providecommand{\doi}[1]{DOI~\discretionary{}{}{}#1}\else
  \providecommand{\doi}{DOI~\discretionary{}{}{}\begingroup \urlstyle{rm}\Url}\fi
\providecommand{\eprint}[2][]{\url{#2}}

\bibitem[{Achiam et~al.(2023)Achiam, Adler, Agarwal, Ahmad, Akkaya, Aleman, Almeida, Altenschmidt, Altman, Anadkat et~al.}]{achiam2023gpt}
Achiam J, Adler S, Agarwal S, Ahmad L, Akkaya I, Aleman FL, Almeida D, Altenschmidt J, Altman S, Anadkat S, et~al. (2023) Gpt-4 technical report. arXiv preprint arXiv:230308774

\bibitem[{Antoniou et~al.(2017)Antoniou, Storkey, and Edwards}]{antoniou2017data}
Antoniou A, Storkey A, Edwards H (2017) Data augmentation generative adversarial networks. arXiv preprint arXiv:171104340

\bibitem[{Bansal and Grover(2023)}]{bansal2023leaving}
Bansal H, Grover A (2023) Leaving reality to imagination: Robust classification via generated datasets. arXiv preprint arXiv:230202503

\bibitem[{BELLEGroup(2023)}]{BELLE}
BELLEGroup (2023) Belle: Be everyone's large language model engine. \url{https://github.com/LianjiaTech/BELLE}

\bibitem[{Bolya and Hoffman(2023)}]{bolya2023token}
Bolya D, Hoffman J (2023) Token merging for fast stable diffusion. In: Proceedings of the IEEE/CVF Conference on Computer Vision and Pattern Recognition, pp 4598--4602

\bibitem[{Bossard et~al.(2014)Bossard, Guillaumin, and Van~Gool}]{bossard2014food}
Bossard L, Guillaumin M, Van~Gool L (2014) Food-101--mining discriminative components with random forests. In: Computer Vision--ECCV 2014: 13th European Conference, Zurich, Switzerland, September 6-12, 2014, Proceedings, Part VI 13, Springer, pp 446--461

\bibitem[{Brock et~al.(2018)Brock, Donahue, and Simonyan}]{brock2018large}
Brock A, Donahue J, Simonyan K (2018) Large scale gan training for high fidelity natural image synthesis. arXiv preprint arXiv:180911096

\bibitem[{Chen et~al.(2022)Chen, Wang, Darrell, and Ebrahimi}]{chen2022contrastive}
Chen D, Wang D, Darrell T, Ebrahimi S (2022) Contrastive test-time adaptation. In: Proceedings of the IEEE/CVF Conference on Computer Vision and Pattern Recognition, pp 295--305

\bibitem[{Chen et~al.(2020)Chen, Kornblith, Norouzi, and Hinton}]{chen2020simple}
Chen T, Kornblith S, Norouzi M, Hinton G (2020) A simple framework for contrastive learning of visual representations. In: International conference on machine learning, PMLR, pp 1597--1607

\bibitem[{Chiang et~al.(2023)Chiang, Li, Lin, Sheng, Wu, Zhang, Zheng, Zhuang, Zhuang, Gonzalez et~al.}]{chiang2023vicuna}
Chiang WL, Li Z, Lin Z, Sheng Y, Wu Z, Zhang H, Zheng L, Zhuang S, Zhuang Y, Gonzalez JE, et~al. (2023) Vicuna: An open-source chatbot impressing gpt-4 with 90\%* chatgpt quality. See https://vicuna lmsys org (accessed 14 April 2023) 2(3):6

\bibitem[{Cimpoi et~al.(2014)Cimpoi, Maji, Kokkinos, Mohamed, and Vedaldi}]{cimpoi2014describing}
Cimpoi M, Maji S, Kokkinos I, Mohamed S, Vedaldi A (2014) Describing textures in the wild. In: Proceedings of the IEEE conference on computer vision and pattern recognition, pp 3606--3613

\bibitem[{Dai et~al.(2023)Dai, Liu, Liao, Huang, Cao, Wu, Zhao, Xu, Liu, Liu et~al.}]{dai2023auggpt}
Dai H, Liu Z, Liao W, Huang X, Cao Y, Wu Z, Zhao L, Xu S, Liu W, Liu N, et~al. (2023) Auggpt: Leveraging chatgpt for text data augmentation. arXiv preprint arXiv:230213007

\bibitem[{Deng et~al.(2009)Deng, Dong, Socher, Li, Li, and Fei-Fei}]{deng2009imagenet}
Deng J, Dong W, Socher R, Li LJ, Li K, Fei-Fei L (2009) Imagenet: A large-scale hierarchical image database. In: 2009 IEEE conference on computer vision and pattern recognition, Ieee, pp 248--255

\bibitem[{Devlin et~al.(2018)Devlin, Chang, Lee, and Toutanova}]{devlin2018bert}
Devlin J, Chang MW, Lee K, Toutanova K (2018) Bert: Pre-training of deep bidirectional transformers for language understanding. arXiv preprint arXiv:181004805

\bibitem[{Fei-Fei et~al.(2004)Fei-Fei, Fergus, and Perona}]{fei2004learning}
Fei-Fei L, Fergus R, Perona P (2004) Learning generative visual models from few training examples: An incremental bayesian approach tested on 101 object categories. In: 2004 conference on computer vision and pattern recognition workshop, IEEE, pp 178--178

\bibitem[{Feng et~al.(2023{\natexlab{a}})Feng, Li, Xu, Liu, Fu, and Zuo}]{feng2023learning}
Feng CM, Li B, Xu X, Liu Y, Fu H, Zuo W (2023{\natexlab{a}}) Learning federated visual prompt in null space for mri reconstruction. In: Proceedings of the IEEE/CVF Conference on Computer Vision and Pattern Recognition, pp 8064--8073

\bibitem[{Feng et~al.(2023{\natexlab{b}})Feng, Yu, Liu, Xu, Khan, and Zuo}]{feng2023towards}
Feng CM, Yu K, Liu N, Xu X, Khan S, Zuo W (2023{\natexlab{b}}) Towards instance-adaptive inference for federated learning. In: Proceedings of the IEEE/CVF International Conference on Computer Vision, pp 23287--23296

\bibitem[{Feng et~al.(2023{\natexlab{c}})Feng, Yu, Liu, Khan, and Zuo}]{feng2023diverse}
Feng CM, Yu K, Liu Y, Khan S, Zuo W (2023{\natexlab{c}}) Diverse data augmentation with diffusions for effective test-time prompt tuning. In: Proceedings of the IEEE/CVF International Conference on Computer Vision, pp 2704--2714

\bibitem[{Gao et~al.(2021)Gao, Geng, Zhang, Ma, Fang, Zhang, Li, and Qiao}]{gao2021clip}
Gao P, Geng S, Zhang R, Ma T, Fang R, Zhang Y, Li H, Qiao Y (2021) Clip-adapter: Better vision-language models with feature adapters. arXiv preprint arXiv:211004544

\bibitem[{Gao et~al.(2024)Gao, Geng, Zhang, Ma, Fang, Zhang, Li, and Qiao}]{gao2024clip}
Gao P, Geng S, Zhang R, Ma T, Fang R, Zhang Y, Li H, Qiao Y (2024) Clip-adapter: Better vision-language models with feature adapters. International Journal of Computer Vision 132(2):581--595

\bibitem[{Gao et~al.(2022)Gao, Shi, Zhu, Wang, Tang, Zhou, Li, and Metaxas}]{gao2022visual}
Gao Y, Shi X, Zhu Y, Wang H, Tang Z, Zhou X, Li M, Metaxas DN (2022) Visual prompt tuning for test-time domain adaptation. arXiv preprint arXiv:221004831

\bibitem[{Goodfellow et~al.(2020)Goodfellow, Pouget-Abadie, Mirza, Xu, Warde-Farley, Ozair, Courville, and Bengio}]{goodfellow2020generative}
Goodfellow I, Pouget-Abadie J, Mirza M, Xu B, Warde-Farley D, Ozair S, Courville A, Bengio Y (2020) Generative adversarial networks. Communications of the ACM 63(11):139--144

\bibitem[{Helber et~al.(2019)Helber, Bischke, Dengel, and Borth}]{helber2019eurosat}
Helber P, Bischke B, Dengel A, Borth D (2019) Eurosat: A novel dataset and deep learning benchmark for land use and land cover classification. IEEE Journal of Selected Topics in Applied Earth Observations and Remote Sensing 12(7):2217--2226

\bibitem[{Hendrycks et~al.(2019)Hendrycks, Mu, Cubuk, Zoph, Gilmer, and Lakshminarayanan}]{hendrycks2019augmix}
Hendrycks D, Mu N, Cubuk ED, Zoph B, Gilmer J, Lakshminarayanan B (2019) Augmix: A simple data processing method to improve robustness and uncertainty. arXiv preprint arXiv:191202781

\bibitem[{Hendrycks et~al.(2021{\natexlab{a}})Hendrycks, Basart, Mu, Kadavath, Wang, Dorundo, Desai, Zhu, Parajuli, Guo et~al.}]{hendrycks2021many}
Hendrycks D, Basart S, Mu N, Kadavath S, Wang F, Dorundo E, Desai R, Zhu T, Parajuli S, Guo M, et~al. (2021{\natexlab{a}}) The many faces of robustness: A critical analysis of out-of-distribution generalization. In: Proceedings of the IEEE/CVF International Conference on Computer Vision, pp 8340--8349

\bibitem[{Hendrycks et~al.(2021{\natexlab{b}})Hendrycks, Zhao, Basart, Steinhardt, and Song}]{hendrycks2021natural}
Hendrycks D, Zhao K, Basart S, Steinhardt J, Song D (2021{\natexlab{b}}) Natural adversarial examples. In: Proceedings of the IEEE/CVF Conference on Computer Vision and Pattern Recognition, pp 15262--15271

\bibitem[{Ho et~al.(2020)Ho, Jain, and Abbeel}]{ho2020denoising}
Ho J, Jain A, Abbeel P (2020) Denoising diffusion probabilistic models. Advances in Neural Information Processing Systems 33:6840--6851

\bibitem[{Ho et~al.(2022)Ho, Chan, Saharia, Whang, Gao, Gritsenko, Kingma, Poole, Norouzi, Fleet et~al.}]{ho2022imagen}
Ho J, Chan W, Saharia C, Whang J, Gao R, Gritsenko A, Kingma DP, Poole B, Norouzi M, Fleet DJ, et~al. (2022) Imagen video: High definition video generation with diffusion models. arXiv preprint arXiv:221002303

\bibitem[{Huang et~al.(2022)Huang, Chu, and Wei}]{huang2022unsupervised}
Huang T, Chu J, Wei F (2022) Unsupervised prompt learning for vision-language models. arXiv preprint arXiv:220403649

\bibitem[{Jia et~al.(2021)Jia, Yang, Xia, Chen, Parekh, Pham, Le, Sung, Li, and Duerig}]{jia2021scaling}
Jia C, Yang Y, Xia Y, Chen YT, Parekh Z, Pham H, Le Q, Sung YH, Li Z, Duerig T (2021) Scaling up visual and vision-language representation learning with noisy text supervision. In: International Conference on Machine Learning, PMLR, pp 4904--4916

\bibitem[{Jia et~al.(2022)Jia, Tang, Chen, Cardie, Belongie, Hariharan, and Lim}]{jia2022visual}
Jia M, Tang L, Chen BC, Cardie C, Belongie S, Hariharan B, Lim SN (2022) Visual prompt tuning. In: Computer Vision--ECCV 2022: 17th European Conference, Tel Aviv, Israel, October 23--27, 2022, Proceedings, Part XXXIII, Springer, pp 709--727

\bibitem[{Karmanov et~al.(2024)Karmanov, Guan, Lu, El~Saddik, and Xing}]{karmanov2024efficient}
Karmanov A, Guan D, Lu S, El~Saddik A, Xing E (2024) Efficient test-time adaptation of vision-language models. In: Proceedings of the IEEE/CVF Conference on Computer Vision and Pattern Recognition, pp 14162--14171

\bibitem[{Kingma and Welling(2013)}]{kingma2013auto}
Kingma DP, Welling M (2013) Auto-encoding variational bayes. arXiv preprint arXiv:13126114

\bibitem[{Krause et~al.(2013)Krause, Stark, Deng, and Fei-Fei}]{krause20133d}
Krause J, Stark M, Deng J, Fei-Fei L (2013) 3d object representations for fine-grained categorization. In: Proceedings of the IEEE international conference on computer vision workshops, pp 554--561

\bibitem[{Le~Scao et~al.(2023)Le~Scao, Fan, Akiki, Pavlick, Ili{\'c}, Hesslow, Castagn{\'e}, Luccioni, Yvon, Gall{\'e} et~al.}]{le2023bloom}
Le~Scao T, Fan A, Akiki C, Pavlick E, Ili{\'c} S, Hesslow D, Castagn{\'e} R, Luccioni AS, Yvon F, Gall{\'e} M, et~al. (2023) Bloom: A 176b-parameter open-access multilingual language model. arXiv preprint arXiv:221105100

\bibitem[{Li et~al.(2022{\natexlab{a}})Li, Feng, Zhou, Xu, and Chang}]{li2022prompt}
Li H, Feng CM, Zhou T, Xu Y, Chang X (2022{\natexlab{a}}) Prompt-driven efficient open-set semi-supervised learning. arXiv preprint arXiv:220914205

\bibitem[{Li et~al.(2022{\natexlab{b}})Li, Zhang, Zhang, Yang, Li, Zhong, Wang, Yuan, Zhang, Hwang et~al.}]{li2022grounded}
Li LH, Zhang P, Zhang H, Yang J, Li C, Zhong Y, Wang L, Yuan L, Zhang L, Hwang JN, et~al. (2022{\natexlab{b}}) Grounded language-image pre-training. In: Proceedings of the IEEE/CVF Conference on Computer Vision and Pattern Recognition, pp 10965--10975

\bibitem[{Liu et~al.(2021)Liu, Kothari, Van~Delft, Bellot-Gurlet, Mordan, and Alahi}]{liu2021ttt++}
Liu Y, Kothari P, Van~Delft B, Bellot-Gurlet B, Mordan T, Alahi A (2021) Ttt++: When does self-supervised test-time training fail or thrive? Advances in Neural Information Processing Systems 34:21808--21820

\bibitem[{Maji et~al.(2013)Maji, Rahtu, Kannala, Blaschko, and Vedaldi}]{maji2013fine}
Maji S, Rahtu E, Kannala J, Blaschko M, Vedaldi A (2013) Fine-grained visual classification of aircraft. arXiv preprint arXiv:13065151

\bibitem[{Mandal et~al.(2019)Mandal, Narayan, Dwivedi, Gupta, Ahmed, Khan, and Shao}]{mandal2019out}
Mandal D, Narayan S, Dwivedi SK, Gupta V, Ahmed S, Khan FS, Shao L (2019) Out-of-distribution detection for generalized zero-shot action recognition. In: Proceedings of the IEEE/CVF Conference on Computer Vision and Pattern Recognition, pp 9985--9993

\bibitem[{Meng et~al.(2023)Meng, Rombach, Gao, Kingma, Ermon, Ho, and Salimans}]{meng2023distillation}
Meng C, Rombach R, Gao R, Kingma D, Ermon S, Ho J, Salimans T (2023) On distillation of guided diffusion models. In: Proceedings of the IEEE/CVF Conference on Computer Vision and Pattern Recognition, pp 14297--14306

\bibitem[{Nichol et~al.(2021)Nichol, Dhariwal, Ramesh, Shyam, Mishkin, McGrew, Sutskever, and Chen}]{nichol2021glide}
Nichol A, Dhariwal P, Ramesh A, Shyam P, Mishkin P, McGrew B, Sutskever I, Chen M (2021) Glide: Towards photorealistic image generation and editing with text-guided diffusion models. arXiv preprint arXiv:211210741

\bibitem[{Nichol and Dhariwal(2021)}]{nichol2021improved}
Nichol AQ, Dhariwal P (2021) Improved denoising diffusion probabilistic models. In: International Conference on Machine Learning, PMLR, pp 8162--8171

\bibitem[{Nilsback and Zisserman(2008)}]{nilsback2008automated}
Nilsback ME, Zisserman A (2008) Automated flower classification over a large number of classes. In: 2008 Sixth Indian Conference on Computer Vision, Graphics \& Image Processing, IEEE, pp 722--729

\bibitem[{Parkhi et~al.(2012)Parkhi, Vedaldi, Zisserman, and Jawahar}]{parkhi2012cats}
Parkhi OM, Vedaldi A, Zisserman A, Jawahar C (2012) Cats and dogs. In: 2012 IEEE conference on computer vision and pattern recognition, IEEE, pp 3498--3505

\bibitem[{Perez and Wang(2017)}]{perez2017effectiveness}
Perez L, Wang J (2017) The effectiveness of data augmentation in image classification using deep learning. arXiv preprint arXiv:171204621

\bibitem[{Piedboeuf and Langlais(2023)}]{piedboeuf2023chatgpt}
Piedboeuf F, Langlais P (2023) Is chatgpt the ultimate data augmentation algorithm? In: Findings of the Association for Computational Linguistics: EMNLP 2023, pp 15606--15615

\bibitem[{Radford et~al.(2018)Radford, Narasimhan, Salimans, Sutskever et~al.}]{radford2018improving}
Radford A, Narasimhan K, Salimans T, Sutskever I, et~al. (2018) Improving language understanding by generative pre-training

\bibitem[{Radford et~al.(2021)Radford, Kim, Hallacy, Ramesh, Goh, Agarwal, Sastry, Askell, Mishkin, Clark et~al.}]{radford2021learning}
Radford A, Kim JW, Hallacy C, Ramesh A, Goh G, Agarwal S, Sastry G, Askell A, Mishkin P, Clark J, et~al. (2021) Learning transferable visual models from natural language supervision. In: International conference on machine learning, PMLR, pp 8748--8763

\bibitem[{Ramesh et~al.(2022)Ramesh, Dhariwal, Nichol, Chu, and Chen}]{ramesh2022hierarchical}
Ramesh A, Dhariwal P, Nichol A, Chu C, Chen M (2022) Hierarchical text-conditional image generation with clip latents. arXiv preprint arXiv:220406125

\bibitem[{Recht et~al.(2019)Recht, Roelofs, Schmidt, and Shankar}]{recht2019imagenet}
Recht B, Roelofs R, Schmidt L, Shankar V (2019) Do imagenet classifiers generalize to imagenet? In: International conference on machine learning, PMLR, pp 5389--5400

\bibitem[{Rombach et~al.(2022)Rombach, Blattmann, Lorenz, Esser, and Ommer}]{rombach2022high}
Rombach R, Blattmann A, Lorenz D, Esser P, Ommer B (2022) High-resolution image synthesis with latent diffusion models. In: Proceedings of the IEEE/CVF Conference on Computer Vision and Pattern Recognition, pp 10684--10695

\bibitem[{Saharia et~al.(2022)Saharia, Chan, Saxena, Li, Whang, Denton, Ghasemipour, Ayan, Mahdavi, Lopes et~al.}]{saharia2022photorealistic}
Saharia C, Chan W, Saxena S, Li L, Whang J, Denton E, Ghasemipour SKS, Ayan BK, Mahdavi SS, Lopes RG, et~al. (2022) Photorealistic text-to-image diffusion models with deep language understanding. arXiv preprint arXiv:220511487

\bibitem[{Schneider et~al.(2020)Schneider, Rusak, Eck, Bringmann, Brendel, and Bethge}]{schneider2020improving}
Schneider S, Rusak E, Eck L, Bringmann O, Brendel W, Bethge M (2020) Improving robustness against common corruptions by covariate shift adaptation. Advances in Neural Information Processing Systems 33:11539--11551

\bibitem[{Shanmugam et~al.(2021)Shanmugam, Blalock, Balakrishnan, and Guttag}]{shanmugam2021better}
Shanmugam D, Blalock D, Balakrishnan G, Guttag J (2021) Better aggregation in test-time augmentation. In: Proceedings of the IEEE/CVF International Conference on Computer Vision, pp 1214--1223

\bibitem[{Shorten and Khoshgoftaar(2019)}]{shorten2019survey}
Shorten C, Khoshgoftaar TM (2019) A survey on image data augmentation for deep learning. Journal of big data 6(1):1--48

\bibitem[{Shu et~al.(2022)Shu, Nie, Huang, Yu, Goldstein, Anandkumar, and Xiao}]{shu2022test}
Shu M, Nie W, Huang DA, Yu Z, Goldstein T, Anandkumar A, Xiao C (2022) Test-time prompt tuning for zero-shot generalization in vision-language models. arXiv preprint arXiv:220907511

\bibitem[{Sinha et~al.(2021)Sinha, Song, Meng, and Ermon}]{sinha2021d2c}
Sinha A, Song J, Meng C, Ermon S (2021) D2c: Diffusion-decoding models for few-shot conditional generation. Advances in Neural Information Processing Systems 34:12533--12548

\bibitem[{Song et~al.(2023)Song, Dhariwal, Chen, and Sutskever}]{song2023consistency}
Song Y, Dhariwal P, Chen M, Sutskever I (2023) Consistency models. arXiv preprint arXiv:230301469

\bibitem[{Soomro et~al.(2012)Soomro, Zamir, and Shah}]{soomro2012ucf101}
Soomro K, Zamir AR, Shah M (2012) Ucf101: A dataset of 101 human actions classes from videos in the wild. arXiv preprint arXiv:12120402

\bibitem[{Sun et~al.(2024)Sun, Zhang, He, Li, Cheng, Liu, Yan, Shao, Tang, Zhang, Zhao, Chen, Zheng, Zhou, Li, Zhan, Zhou, Li, Yang, Wu, Yin, Huang, Jiang, and Qiu}]{Sun2024MOSS}
Sun T, Zhang X, He Z, Li P, Cheng Q, Liu X, Yan H, Shao Y, Tang Q, Zhang S, Zhao X, Chen K, Zheng Y, Zhou Z, Li R, Zhan J, Zhou Y, Li L, Yang X, Wu L, Yin Z, Huang X, Jiang YG, Qiu X (2024) Moss: An open conversational large language model. Machine Intelligence Research \urlprefix\url{https://github.com/OpenMOSS/MOSS}

\bibitem[{Sun et~al.(2020)Sun, Wang, Liu, Miller, Efros, and Hardt}]{sun2020test}
Sun Y, Wang X, Liu Z, Miller J, Efros A, Hardt M (2020) Test-time training with self-supervision for generalization under distribution shifts. In: International conference on machine learning, PMLR, pp 9229--9248

\bibitem[{Touvron et~al.(2023)Touvron, Lavril, Izacard, Martinet, Lachaux, Lacroix, Rozi{\`e}re, Goyal, Hambro, Azhar et~al.}]{touvron2023llama}
Touvron H, Lavril T, Izacard G, Martinet X, Lachaux MA, Lacroix T, Rozi{\`e}re B, Goyal N, Hambro E, Azhar F, et~al. (2023) Llama: Open and efficient foundation language models. arXiv preprint arXiv:230213971

\bibitem[{Ubani et~al.(2023)Ubani, Polat, and Nielsen}]{ubani2023zeroshotdataaug}
Ubani S, Polat SO, Nielsen R (2023) Zeroshotdataaug: Generating and augmenting training data with chatgpt. arXiv preprint arXiv:230414334

\bibitem[{Wang et~al.(2020)Wang, Shelhamer, Liu, Olshausen, and Darrell}]{wang2020tent}
Wang D, Shelhamer E, Liu S, Olshausen B, Darrell T (2020) Tent: Fully test-time adaptation by entropy minimization. arXiv preprint arXiv:200610726

\bibitem[{Wang et~al.(2019)Wang, Ge, Lipton, and Xing}]{wang2019learning}
Wang H, Ge S, Lipton Z, Xing EP (2019) Learning robust global representations by penalizing local predictive power. Advances in Neural Information Processing Systems 32

\bibitem[{Xiao et~al.(2010)Xiao, Hays, Ehinger, Oliva, and Torralba}]{xiao2010sun}
Xiao J, Hays J, Ehinger KA, Oliva A, Torralba A (2010) Sun database: Large-scale scene recognition from abbey to zoo. In: 2010 IEEE computer society conference on computer vision and pattern recognition, IEEE, pp 3485--3492

\bibitem[{Zhang et~al.(2020)Zhang, Deng, Kawaguchi, Ghorbani, and Zou}]{zhang2020does}
Zhang L, Deng Z, Kawaguchi K, Ghorbani A, Zou J (2020) How does mixup help with robustness and generalization? arXiv preprint arXiv:201004819

\bibitem[{Zhang et~al.(2021{\natexlab{a}})Zhang, Levine, and Finn}]{zhang2021memo}
Zhang M, Levine S, Finn C (2021{\natexlab{a}}) Memo: Test time robustness via adaptation and augmentation. arXiv preprint arXiv:211009506

\bibitem[{Zhang et~al.(2022{\natexlab{a}})Zhang, Cai, Pan, Hong, Guo, Yang, and Liu}]{zhang2022motiondiffuse}
Zhang M, Cai Z, Pan L, Hong F, Guo X, Yang L, Liu Z (2022{\natexlab{a}}) Motiondiffuse: Text-driven human motion generation with diffusion model. arXiv preprint arXiv:220815001

\bibitem[{Zhang et~al.(2021{\natexlab{b}})Zhang, Fang, Zhang, Gao, Li, Dai, Qiao, and Li}]{zhang2021tip}
Zhang R, Fang R, Zhang W, Gao P, Li K, Dai J, Qiao Y, Li H (2021{\natexlab{b}}) Tip-adapter: Training-free clip-adapter for better vision-language modeling. arXiv preprint arXiv:211103930

\bibitem[{Zhang et~al.(2022{\natexlab{b}})Zhang, Wang, Zhou, Schuurmans, and Gonzalez}]{zhang2022tempera}
Zhang T, Wang X, Zhou D, Schuurmans D, Gonzalez JE (2022{\natexlab{b}}) Tempera: Test-time prompting via reinforcement learning. arXiv preprint arXiv:221111890

\bibitem[{Zhang et~al.(2023)Zhang, Wei, Jiang, Zhang, Zuo, and Tian}]{zhang2023controlvideo}
Zhang Y, Wei Y, Jiang D, Zhang X, Zuo W, Tian Q (2023) Controlvideo: Training-free controllable text-to-video generation. arXiv preprint arXiv:230513077

\bibitem[{Zhao et~al.(2020)Zhao, Liu, Lin, Zhu, and Han}]{zhao2020differentiable}
Zhao S, Liu Z, Lin J, Zhu JY, Han S (2020) Differentiable augmentation for data-efficient gan training. Advances in Neural Information Processing Systems 33:7559--7570

\bibitem[{Zhong et~al.(2020)Zhong, Zheng, Kang, Li, and Yang}]{zhong2020random}
Zhong Z, Zheng L, Kang G, Li S, Yang Y (2020) Random erasing data augmentation. In: Proceedings of the AAAI conference on artificial intelligence, vol~34, pp 13001--13008

\bibitem[{Zhou et~al.(2022{\natexlab{a}})Zhou, Yang, Loy, and Liu}]{zhou2022conditional}
Zhou K, Yang J, Loy CC, Liu Z (2022{\natexlab{a}}) Conditional prompt learning for vision-language models. In: Proceedings of the IEEE/CVF Conference on Computer Vision and Pattern Recognition, pp 16816--16825

\bibitem[{Zhou et~al.(2022{\natexlab{b}})Zhou, Yang, Loy, and Liu}]{zhou2022learning}
Zhou K, Yang J, Loy CC, Liu Z (2022{\natexlab{b}}) Learning to prompt for vision-language models. International Journal of Computer Vision 130(9):2337--2348

\end{thebibliography}


\documentclass{article}
\begin{document}
    \nocite{*}
    \bibliography{sn-bibliography} 
    \bibliographystyle{sn-basic} 
\end{document}

%
%
%


\end{sloppypar}

\end{document}


\begin{sloppypar}
\title{Noise-Robust Federated Learning with Heterogeneous Clients
}

\author{Anonymous Authors}
\author{Xiuwen Fang         \and
Mang Ye
}
%
%
\institute{Xiuwen Fang and Mang Ye \at
    National Engineering Research Center for Multimedia Software, Institute of Artificial Intelligence,\\ Hubei Key Laboratory of Multimedia and Network Communication Engineering, \\
    School of Computer Science, Wuhan University, Wuhan, China \\
              \email{fangxiuwen, yemang@whu.edu.cn}           
}

\date{Received: date / Accepted: date}

\maketitle

\begin{appendix}

\definecolor{mygray}{gray}{.9}
\begin{table}[b]\small
\centering
 \caption{\label{tab:parazeta} Parametric analysis for $\zeta$ in the DLR module, when the noise rate is $0.1$ and the noise type is pairflip.
 }
\setlength{\tabcolsep}{3.1mm}{
\begin{tabular}{c||cccc|c}\hline
  \specialrule{.1em}{0em}{0em}
\rowcolor{mygray}
  & \multicolumn{5}{c}{Pairflip, $\mu=0.1$}  \\\hhline{~||-----}
\rowcolor{mygray} \multirow{-2}{*}{$\zeta$} &$\theta_1$ &$\theta_2$ &$\theta_3$ &$\theta_4$ &Avg   \\\hhline{-||-----}
0.1&	75.61&	72.26&	62.41&	64.71&	68.75 \\
0.5&	80.25& 	79.29&	71.07&	77.12&	76.93 \\
1.0&	79.37&	78.94&	71.82&	\textbf{78.83}&	77.24 \\
4.0&	78.35& 	80.05&	\textbf{72.59}&	78.18&	77.29 \\
\textbf{10.0}&	\textbf{80.32}&	\textbf{81.12}& 	71.71& 	78.02& 	\textbf{77.79} \\
12.0&	80.16& 	79.83&	72.53&	77.30&	77.46 \\
14.0&	79.69&	79.35&	71.65&	77.11&	76.95 \\
100.0&	78.98&	79.61&	70.44&	76.69&	76.43 \\\hline
  \specialrule{.1em}{0em}{0em}
 \end{tabular}}
  \vspace{-2mm}
\end{table}

\definecolor{mygray}{gray}{.9}
\begin{table}[b]\small
\centering
 \caption{\label{tab:paralamda} Parametric analysis for $\lambda$ in the SL module, when the noise rate is $0.1$ and the noise type is pairflip.
 }
\setlength{\tabcolsep}{3.1mm}{
\begin{tabular}{c|c||cccc|c}\hline
  \specialrule{.1em}{0em}{0em}
\rowcolor{mygray}
 &  & \multicolumn{5}{c}{Pairflip, $\mu=0.1$}  \\\hhline{~|~||-----}
\rowcolor{mygray} \multirow{-2}{*}{$\lambda$} & \multirow{-2}{*}{$\gamma$} &$\theta_1$ &$\theta_2$ &$\theta_3$ &$\theta_4$ &Avg   \\\hhline{-|-||-----}
0.1&	 &	75.91&	75.23&	69.86&	76.16&	74.29 \\
0.2&	&	\textbf{80.44}&	78.84&	71.13&	77.52&	76.98 \\
0.3&		&79.29&	80.10&	\textbf{72.01}&	77.68&	77.27 \\
\textbf{0.4}&		&80.32& 	\textbf{81.12}& 	71.71& 	\textbf{78.02}& 	\textbf{77.79} \\
0.5&	\multirow{-5}{*}{0.9}&	79.88&	79.70&	71.75&	77.49&	77.21 \\\hline
  \specialrule{.1em}{0em}{0em}
 \end{tabular}}
  \vspace{-2mm}
\end{table}

\definecolor{mygray}{gray}{.9}
\begin{table}[b]\small
\centering
 \caption{\label{tab:paragamma} Parametric analysis for $\gamma$ in the SL module, when the noise rate is $0.1$ and the noise type is pairflip.
 }
\setlength{\tabcolsep}{3.1mm}{
\begin{tabular}{c|c||cccc|c}\hline
  \specialrule{.1em}{0em}{0em}
\rowcolor{mygray}
 &  & \multicolumn{5}{c}{Pairflip, $\mu=0.1$}  \\\hhline{~|~||-----}
\rowcolor{mygray} \multirow{-2}{*}{$\lambda$} & \multirow{-2}{*}{$\gamma$} &$\theta_1$ &$\theta_2$ &$\theta_3$ &$\theta_4$ &Avg   \\\hhline{-|-||-----}
	&0.4&	78.97&	78.83&	72.02&	77.95&	76.94 \\
	&0.5&	79.65&	78.89&	\textbf{72.69}&	78.10&	77.33 \\
	&0.6&	79.84&	80.14&	72.00&	\textbf{78.69}&	77.67 \\
	&0.7&	79.44&	78.43&	72.64&	77.03&	76.89 \\
	&0.8&	79.37&	80.22&	70.41&	77.67&	76.92 \\
	&\textbf{0.9}&	\textbf{80.32}& 	\textbf{81.12}& 	71.71& 	78.02& 	\textbf{77.79} \\
\multirow{-7}{*}{0.4}	&1.0&	79.12&	79.70&	71.68&	78.57&	77.27 \\\hline
  \specialrule{.1em}{0em}{0em}
 \end{tabular}}
  \vspace{-2mm}
\end{table}

\definecolor{mygray}{gray}{.9}
\begin{table}[b]\small
\centering
 \caption{\label{tab:paratau} Parametric analysis for temperature hyperparameter $\tau$ in the SL module, when the noise rate is $0.1$ and the noise type is pairflip.
 }
\setlength{\tabcolsep}{3.1mm}{
\begin{tabular}{c||cccc|c}\hline
  \specialrule{.1em}{0em}{0em}
\rowcolor{mygray}
  & \multicolumn{5}{c}{Pairflip, $\mu=0.1$}  \\\hhline{~||-----}
\rowcolor{mygray} \multirow{-2}{*}{$\tau$} &$\theta_1$ &$\theta_2$ &$\theta_3$ &$\theta_4$ &Avg   \\\hhline{-||-----}
0.5&	79.97&	78.90&	70.06&	70.06&	74.75 \\
2.0&	79.82&	78.33&	71.56&	77.00&	76.68 \\
\textbf{4.0}&	80.32& 	\textbf{81.12}& 	71.71& 	78.02& 	\textbf{77.79} \\
6.0&	\textbf{80.59}&	78.51&	71.77&	78.15&	77.26 \\
8.0&	78.31&	79.54&	\textbf{72.51}&	\textbf{78.93}&	77.32 \\
10.0&	78.49&	80.22&	71.38&	75.88&	76.49 \\\hline
  \specialrule{.1em}{0em}{0em}
 \end{tabular}}
  \vspace{-2mm}
\end{table}

\section{Ablation Study on Hyper-Parameters}
We conduct additional experiments to investigate the impact of different values of $\zeta$ in Eq.5, $\lambda$ and $\gamma$ in Eq.9, and $\tau$ in Eq.10 on our algorithm. 1) $\zeta$ is a hyperparameter that controls the maximum dependence on the model prediction in DLR. We test $\zeta\in[0.1,100]$ on Cifar10 with a noise rate of $0.1$ and a noise type of pairflip. From Table~\ref{tab:parazeta}, we can find that smaller $\zeta$ often causes the soft labels calculated by DLR to be overly dependent on model predictions. Excessively large $\zeta$ will cause the soft label to be overly dependent on the original noise label, weakening the effect of DLR in resisting noise. Therefore, we chose a trade-off optimal value of $\zeta=10.0$. 
2) $\lambda$ and $\gamma$ are a pair of combined hyperparameters that control the SL loss function, where $\lambda$ controls the overfitting of CE to noise and $\gamma$ controls the effect of RCE. We test $\lambda\in[0.1,0.5]$, $\gamma=0.9$ and $\lambda=0.4$, $\gamma\in[0.4,1.0]$ on Cifar10 with a noise rate of $0.1$ and a noise type of pairflip. From Table~\ref{tab:paralamda}, we can observe that larger $\lambda$ leads to more overfitting, while smaller $\lambda$ can help alleviate the overfitting of CE. However, when $\lambda$ is too small, the SL function approaches RCE alone and the convergence may become slow. Therefore, a relatively large $\lambda$ can help convergence on difficult datasets. From Table~\ref{tab:paragamma}, we can observe that a larger $\gamma$ can alleviate the fitting to the original noisy label during model training. However, $\gamma$ needs to be adjusted in proportion with $\lambda$, and should not be too large. Therefore, we chose a set of trade-off optimal values $\lambda=0.4$, $\gamma=0.9$. 
3) The temperature hyperparameter $\tau$ is utilized to control the peakiness of the output probability distribution. We test $\tau\in[0.5,10]$ on Cifar10 with a noise rate of $0.1$ and a noise type of pairflip, as shown in Table~\ref{tab:paratau}. With a larger temperature, the softmax function produces a flatter probability distribution, which prevents the model from being overconfident and thus improves the ability of the model to handle unseen data during inference. A higher temperature also helps to regularize the model and prevents it from memorizing noisy labels in the training data. However, an excessively high temperature will blindly soften the output distribution, so we set $\tau$ to a reasonable value of $4.0$.

The results show that our algorithm is robust and stable under a wide range of hyper-parameter values, and that there exists an optimal value for each hyper-parameter that balances between different objectives and constraints. The sensitivity and adaptability of our algorithm in different federated learning settings are demonstrated.

\begin{figure*}[t]
\centering{
    \includegraphics[width=0.33\linewidth]{figs/0.5noniid.pdf}
    \includegraphics[width=0.33\linewidth]{figs/1.0noniid.pdf}
    \includegraphics[width=0.33\linewidth]{figs/5.0noniid.pdf} 
    \caption{An illustration of the number of samples for each class in each client (expressed as point sizes), when the Dirichlet distribution beta value is 0.5, 1.0 and 5.0 (from left to right)}
    \label{fig:noniidbeta}}
\end{figure*}

\definecolor{mygray}{gray}{.9}
\begin{table*}[t]\small
\centering
 \caption{\label{tab:noniid1} Using Cifar10 as the private dataset and Cifar100 as the public dataset. Comparison with the state-of-the-art method in data heterogeneous scenario when the noise rate $\mu=0.1$ and $\theta_k$ represents the local model of the client $c_k$.
 }
\setlength{\tabcolsep}{3.1mm}{
\begin{tabular}{c||c||cccc|c||cccc|c}\hline
  \specialrule{.1em}{0em}{0em}
\rowcolor{mygray}
 &  & \multicolumn{5}{c||}{Pairflip}  & \multicolumn{5}{c}{Symflip}\\\hhline{~||~||-----||-----}
\rowcolor{mygray}  \multirow{-2}{*}{beta} & \multirow{-2}{*}{Method} &$\theta_1$ &$\theta_2$ &$\theta_3$ &$\theta_4$ &Avg    &$\theta_1$ &$\theta_2$ &$\theta_3$ &$\theta_4$ &Avg  \\\hhline{-||-||-----||-----}
 &Baseline &62.84	&54.05	&38.88	&49.83	&51.40	&58.70	&55.18	&37.74	&47.65	&49.82  \\
 &FedMD &66.88	&61.06	&46.53	&54.50	&57.24	&64.79	&63.35	&42.18	&57.09	&55.94  \\
 &FedDF &70.78	&61.12	&45.04	&57.90	&58.71	&64.63	&61.22	&43.63	&54.26	&56.52  \\
 &KT-pFL &70.42	&68.94	&43.46	&57.49	&60.08	&69.24 	&66.25	&43.69	&54.34	&58.38\\
 &FedProto &69.67	&67.90	&45.28	&56.65	&59.88	&68.86	&66.14	&43.23	&54.21	&58.11\\
 &FCCL &70.29	&65.72	&44.02	&55.69	&58.93	&66.34	&63.54	&42.82	&53.36	&57.43 \\\hhline{~||-||-----||-----}
 &RHFL &70.81	&67.56	&41.55	&49.76	&57.42	&71.48	&66.55	&43.04	&48.63	&56.38 \\
 \multirow{-8}{*}{0.5}&\textbf{RHFL+} &70.01	&67.43	&45.36	&50.69	&58.37	&71.73	&67.60	&43.36	&51.08	&58.44 \\
 \hline
 &Baseline &58.48	&65.56	&57.89	&59.42	&60.34	&56.30	&64.19	&57.04	&55.58	&58.28 \\
 &FedMD &60.85	&68.29	&67.35	&64.00	&65.12	&60.06	&65.16	&65.15	&61.64	&63.00  \\
 &FedDF &59.91	&68.25	&64.58	&61.61	&63.59	&60.61	&63.70	&64.10	&61.35	&62.44  \\
 &KT-pFL &63.81 	&73.51	&65.51	&63.94	&66.69	&63.32	&70.04	&63.06	&61.80	&64.56\\
 &FedProto &63.62 	&72.92	&64.78	&63.87	&66.30	&63.23	&69.89	&62.56	&60.88	&64.14\\
 &FCCL &63.60	&71.23	&67.33	&65.38	&66.89	&61.85	&69.46	&65.12	&62.27	&64.68  \\\hhline{~||-||-----||-----}
 &RHFL &62.88	&73.42	&66.13	&60.27	&65.68	&64.31	&73.84	&67.42	&60.33	&66.48 \\
 \multirow{-8}{*}{1.0}&\textbf{RHFL+} &63.74	&76.04	&70.05	&62.32	&68.04	&64.04	&72.18	&67.52	&61.13	&66.22 \\
 \hline
 &Baseline &78.38	&67.38	&61.61	&76.46	&70.96	&76.30	&65.82	&60.56	&74.43	&69.28 \\
 &FedMD &76.14	&66.94	&64.87	&75.97	&70.98	&74.99	&67.25	&63.70	&75.79	&70.43   \\
 &FedDF &75.07	&67.94	&63.94	&77.19	&71.04	&72.58	&65.82	&63.22	&75.06	&69.17   \\
 &KT-pFL &79.81	&72.40	&63.29	&79.77	&73.82	&77.71	&70.31	&63.17	&77.62	&72.20 \\
 &FedProto &79.56	&72.17	&63.76	&79.51	&73.75	&78.67	&70.86	&63.35	&77.12	&72.50 \\
 &FCCL &79.93	&71.70	&67.67	&80.10	&74.85	&79.09	&70.11	&66.64	&78.80	&73.66 \\\hhline{~||-||-----||-----}
 &RHFL &79.77	&72.78	&68.23	&78.57	&74.84	&80.70	&70.21	&67.14	&78.23	&74.07 \\
 \multirow{-8}{*}{5.0}&\textbf{RHFL+} &80.03	&71.44	&70.41	&80.37	&75.56	&80.16	&73.09	&70.09	&80.13	&75.87 \\
 \hline
  \specialrule{.1em}{0em}{0em}
 \end{tabular}}
  \vspace{-2mm}
\end{table*}

\definecolor{mygray}{gray}{.9}
\begin{table*}[t]\small
\centering
 \caption{\label{tab:noniid2} Using Cifar10 as the private dataset and Cifar100 as the public dataset. Comparison with the state-of-the-art method in data heterogeneous scenario when the noise rate $\mu=0.1$ and $\theta_k$ represents the local model of the client $c_k$.
 }
\setlength{\tabcolsep}{3.1mm}{
\begin{tabular}{c||c||cccc|c||cccc|c}\hline
  \specialrule{.1em}{0em}{0em}
\rowcolor{mygray}
 &  & \multicolumn{5}{c||}{Pairflip}  & \multicolumn{5}{c}{Symflip}\\\hhline{~||~||-----||-----}
\rowcolor{mygray}  \multirow{-2}{*}{beta} & \multirow{-2}{*}{Method} &$\theta_1$ &$\theta_2$ &$\theta_3$ &$\theta_4$ &Avg    &$\theta_1$ &$\theta_2$ &$\theta_3$ &$\theta_4$ &Avg  \\\hhline{-||-||-----||-----}
 &Baseline &58.32	&53.83	&38.48	&49.27	&49.98	&57.71 	&52.70	&35.78	&47.05	&48.31  \\
 &FedMD &57.50	&56.95	&43.24	&54.92	&53.15	&57.34	&58.59	&40.03	&48.31	&51.07  \\
 &FedDF &62.47	&54.80	&43.73	&55.41	&54.10	&54.62	&56.23	&41.86	&48.97	&50.42  \\
 &KT-pFL &66.02 	&62.02	&38.80	&54.57	&55.35	&64.13 	&63.55	&39.66	&49.84	&54.30 \\
 &FedProto &65.33	&61.60	&41.95	&54.23	&55.78	&63.93	&62.81	&38.48	&49.42	&53.66 \\
 &FCCL &63.51	&64.68	&43.08	&53.44	&56.18	&65.45	&62.89	&41.55	&49.52	&54.85  \\\hhline{~||-||-----||-----}
 &RHFL &66.74	&58.22	&41.81	&49.02	&53.95	&68.10	&63.84	&42.66	&48.01	&55.65 \\
 \multirow{-8}{*}{0.5}&\textbf{RHFL+} &66.63	&63.50	&43.64	&50.30	&56.02	&67.41	&65.39	&43.05	&49.35	&56.30 \\\hline
 &Baseline &50.81	&58.87	&5.92	&58.15	&55.94	&53.19	&62.58	&57.26	&53.89	&56.73  \\
 &FedMD &55.21	&64.40	&58.74	&62.04	&60.10	&53.42	&59.78	&60.93	&58.00	&58.03  \\
 &FedDF &54.58	&60.32	&61.26	&60.35	&59.13	&53.77	&62.56	&61.19	&56.33	&58.46  \\
 &KT-pFL &60.06	&66.14	&59.89	&63.47	&62.39	&59.36 	&66.20	&59.27	&57.91	&60.69 \\
 &FedProto &59.76	&66.28	&60.12	&62.89	&62.26	&58.93	&66.26	&59.07	&57.24	&60.38\\
 &FCCL &59.75	&65.14	&64.42	&63.23	&63.14	&59.68	&66.24	&64.10	&59.43	&62.36 \\\hhline{~||-||-----||-----}
 &RHFL &61.98	&68.57	&62.91	&57.65	&62.78	&63.21	&69.04	&63.20	&57.58	&63.26 \\
 \multirow{-8}{*}{1.0}&\textbf{RHFL+} &62.23	&71.38	&66.55	&60.23	&65.10	&61.13	&72.75	&66.83	&60.07	&65.20 \\\hline
 &Baseline &69.85	&63.53	&56.43	&74.01	&65.96	&70.59	&62.68	&57.37	&70.33	&65.24  \\
 &FedMD &71.09	&62.39	&59.93	&73.81	&66.81	&67.61	&62.62	&60.34	&68.82	&64.85  \\
 &FedDF &68.65	&62.62	&62.40	&74.43	&67.03	&65.37	&62.87	&59.19	&69.84	&64.32  \\
 &KT-pFL &73.26	&68.60	&56.07	&77.28	&68.80	&71.74 	&66.30	&56.85	&74.02	&67.23 \\
 &FedProto &72.88	&68.06	&57.07	&77.45	&68.87	&71.93	&66.35	&57.80	&73.47	&67.39 \\
 &FCCL &73.25	&69.28	&63.21	&78.99	&71.18	&71.98	&67.21	&62.37	&76.00	&69.39 \\\hhline{~||-||-----||-----}
 &RHFL &75.23	&71.54	&66.26	&77.07	&72.53	&78.66	&70.13	&65.38	&77.61	&72.95 \\
 \multirow{-8}{*}{5.0}&\textbf{RHFL+} &74.99	&66.84	&65.27	&78.89	&71.50	&76.96	&70.08	&68.82	&77.33	&73.30 \\\hline
  \specialrule{.1em}{0em}{0em}
 \end{tabular}}
  \vspace{-2mm}
\end{table*}

\section{Test in Data Heterogeneous Scenarios}
To verify the effectiveness of RHFL+ in various complex federated learning scenarios, we evaluated our algorithm under different levels of heterogeneous data. In the Non-IID setting, the label distribution of each client follows a Dirichlet distribution. For each client, samples are sampled from the whole training dataset according to the label distribution following the Dirichlet distribution. The hyperparameter beta of the Dirichlet distribution controls the degree of data heterogeneity (the smaller the beta, the higher the degree of data heterogeneity). We set it to $0.5$, $1.0$ and $5.0$ to investigate the effect of different levels of data heterogeneity on our algorithm. Figure~\ref{fig:noniidbeta} visualizes the sample distribution of each client when beta=$0.5$, $1.0$, and $5.0$. \textcolor{blue}{In order to adapt to the new experimental settings, we set the temperature parameter $\tau$ to $2.0$, the $\eta$ in ECCR that controls the impact of the client confidence indicator is set to $2.0$, and other parameters are consistent with those in the Section 4.1.}

\textcolor{blue}{The corresponding results are shown in Tables~\ref{tab:noniid1}\&\ref{tab:noniid2}. The results show that although our algorithm does not design a corresponding solution module for the data heterogeneity problem, it still shows adaptability in different levels of data heterogeneity.
When the Dirichlet distribution beta values are 1.0 and 5.0, our RHFL+ algorithm shows good competitiveness compared with the SOTA method. However, the accuracy of RHFL+ seems to be limited when the Dirichlet distribution beta value is 0.5 and the noise rate is 0.1. The data heterogeneous scenario with beta of 0.5 can be regarded as an extreme data heterogeneity situation, and the effect of RHFL+ is hindered. In addition, RHFL+ has good noise robustness as it performs better at high noise rates.
Our results indicate that while RHFL+ performs relatively well in handling mild to moderate data heterogeneity, its efficacy may be limited in scenarios of extreme data heterogeneity. We acknowledge this limitation and will consider the diversity and complexity of data patterns among clients in future work, enabling the method to be robustly applied to federated learning scenarios with extreme data heterogeneity. }

\definecolor{mygray}{gray}{.9}
\begin{table}[t]\small
\centering
 \caption{\label{tab:numclient_hetero_cifar10} \textcolor{blue}{Using Cifar10 as the private dataset and CelabA as the public dataset. Comparison with state-of-the-art methods in various noisy heterogeneous scenarios. $\mu$ represents the noise rate. The presented values depict the average accuracy across all clients. }
 }
\setlength{\tabcolsep}{3.1mm}{
\begin{tabular}{c||cc|cc}\hline
  \specialrule{.1em}{0em}{0em}
\rowcolor{mygray}
  & \multicolumn{2}{c|}{Pairflip}& \multicolumn{2}{c}{Symflip}\\\hhline{~||--|--}
\rowcolor{mygray} \multirow{-2}{*}{Method} &$\mu=0.1$ &$\mu=0.2$ &$\mu=0.1$ &$\mu=0.2$   \\\hhline{-||--|--}
Baseline	&39.88	&36.21	&40.47	&35.92 \\
FedMD	&42.65	&37.81	&42.25	&37.52 \\
FedDF	&43.02	&37.89	&42.51	&36.88 \\
KT-pFL	&44.17	&38.81	&44.36	&39.38 \\
FedProto	&44.82	&39.99	&44.63	&39.97 \\
FCCL	&45.29	&39.87	&44.72	&40.03 \\\hline
RHFL	&45.91	&39.20	&45.31	&41.16 \\
\textbf{RHFL+}	&44.97	&40.60	&45.64	&42.18 \\\hline
  \specialrule{.1em}{0em}{0em}
 \end{tabular}}
  \vspace{-2mm}
\end{table}

\section{Test in FL Scenarios with Numerous Clients}

\definecolor{mygray}{gray}{.9}
\begin{table}[t]\small
\centering
 \caption{\label{tab:numclient_hetero_cifar100} \textcolor{blue}{Using Cifar100 as the private dataset and CelabA as the public dataset. Comparison with state-of-the-art methods in various noisy model heterogeneous scenarios. $\mu$ represents the noise rate. The presented values depict the average accuracy across all clients. }
 }
\setlength{\tabcolsep}{3.1mm}{
\begin{tabular}{c||cc|cc}\hline
  \specialrule{.1em}{0em}{0em}
\rowcolor{mygray}
  & \multicolumn{2}{c|}{Pairflip}& \multicolumn{2}{c}{Symflip}\\\hhline{~||--|--}
\rowcolor{mygray} \multirow{-2}{*}{Method} &$\mu=0.1$ &$\mu=0.2$ &$\mu=0.1$ &$\mu=0.2$   \\\hhline{-||--|--}
Baseline	&10.96	&9.62	&10.96	&9.62 \\
FedMD	&11.37	&9.98	&11.41	&9.85 \\
FedDF	&11.32	&10.06	&11.36	&9.66 \\
KT-pFL	&12.37	&10.81	&12.32	&10.72 \\
FedProto	&12.48	&10.96	&12.42	&10.90 \\
FCCL	&12.12	&10.46	&12.02	&10.41 \\\hline
RHFL	&11.91	&10.42	&11.97	&10.81 \\
\textbf{RHFL+}	&12.38	&10.89	&12.50	&11.07\\\hline
  \specialrule{.1em}{0em}{0em}
 \end{tabular}}
  \vspace{-2mm}
\end{table}

\textcolor{blue}{To further verify the applicability of our method in real-world scenarios, we conduct unified experiments while using Cifar10 or Cifar100 as the local dataset in a federated learning scenario containing $40$ model heterogeneous clients, as shown in Table~\ref{tab:numclient_hetero_cifar10}\&\ref{tab:numclient_hetero_cifar100}. We introduce the problem of label shift between clients while studying label noise, where the label distribution of each client follows the Dirichlet distribution. We set the hyperparameter beta of the Dirichlet distribution to $5.0$. In order to adapt to the experimental settings, we set the temperature parameter $\tau$ to $2.0$, the $\eta$ in ECCR that controls the impact of the client confidence indicator is set to $2.0$, , and other parameters are consistent with those in the Section 4.1. Due to the very large number of clients involved, the dataset is randomly distributed among clients without overlap, resulting in a small number of local samples per client. The local models demonstrate swift convergence owing to the smaller private datasets available to each client. Consequently, for this experiment, we decrease the training epochs $T_c$ to accommodate these dynamics. }

\textcolor{blue}{From the results, we can see that even in the federated learning scenarios with label shift containing a large number of clients, our method achieves significant results compared to other SOTA methods. Furthermore, our method shows higher accuracy on the Cifar10 dataset (Tabel~\ref{tab:numclient_hetero_cifar10}) compared to the Cifar100 (Table~\ref{tab:numclient_hetero_cifar100}) dataset. Because we use random sampling without replacement to allocate local samples to clients. When using the Cifar100 dataset, clients often lack samples for multiple categories, resulting in extremely high data heterogeneity in the local dataset.}

\definecolor{mygray}{gray}{.9}
\begin{table}[t]\small
\centering
 \caption{\label{tab:numclient_homo_cifar10} \textcolor{blue}{Using Cifar10 as the private dataset and CelabA as the public dataset. Comparison with state-of-the-art methods in various noisy model homogeneous scenarios. $\mu$ represents the noise rate. The presented values depict the average accuracy across all clients. }
 }
\setlength{\tabcolsep}{3.1mm}{
\begin{tabular}{c||cc|cc}\hline
  \specialrule{.1em}{0em}{0em}
\rowcolor{mygray}
  & \multicolumn{2}{c|}{Pairflip}& \multicolumn{2}{c}{Symflip}\\\hhline{~||--|--}
\rowcolor{mygray} \multirow{-2}{*}{Method} &$\mu=0.1$ &$\mu=0.2$ &$\mu=0.1$ &$\mu=0.2$   \\\hhline{-||--|--}
Baseline	&41.95	&36.51	&41.57	&37.00 \\
FedMD	&42.70	&38.39	&43.74	&38.44 \\
FedDF	&44.14	&38.95	&44.15	&38.15 \\
KT-pFL	&46.60	&41.31	&46.84	&41.54 \\
FedProto	&47.47	&42.13	&47.65	&42.35 \\
FCCL	&45.39	&39.97	&46.53	&40.73 \\\hline
RHFL	&46.03	&40.69	&47.74	&43.33 \\
\textbf{RHFL+}	&47.19	&42.71	&47.88	&44.00 \\\hline
  \specialrule{.1em}{0em}{0em}
 \end{tabular}}
  \vspace{-2mm}
\end{table}

\definecolor{mygray}{gray}{.9}
\begin{table}[t]\small
\centering
 \caption{\label{tab:numclient_homo_cifar100} \textcolor{blue}{Using Cifar100 as the private dataset and CelabA as the public dataset. Comparison with state-of-the-art methods in various noisy model homogeneous scenarios. $μ$ represents the noise rate. The presented values depict the average accuracy across all clients. }
 }
\setlength{\tabcolsep}{3.1mm}{
\begin{tabular}{c||cc|cc}\hline
  \specialrule{.1em}{0em}{0em}
\rowcolor{mygray}
  & \multicolumn{2}{c|}{Pairflip}& \multicolumn{2}{c}{Symflip}\\\hhline{~||--|--}
\rowcolor{mygray} \multirow{-2}{*}{Method} &$\mu=0.1$ &$\mu=0.2$ &$\mu=0.1$ &$\mu=0.2$   \\\hhline{-||--|--}
Baseline	&11.82	&10.44	&11.70	&10.31 \\
FedMD	&11.73	&10.25	&11.56	&9.91 \\
FedDF	11.89	&10.17	&11.69	&10.02 \\
KT-pFL	&13.09	&11.46	&13.05	&11.25 \\
FedProto	&13.32	&11.61	&13.23	&11.51 \\
FCCL	&12.60	&11.07	&12.62	&10.99 \\\hline
RHFL	&12.45	&10.84	&12.58	&11.10 \\
\textbf{RHFL+}	&13.05	&11.43	&13.22	&11.54 \\\hline
  \specialrule{.1em}{0em}{0em}
 \end{tabular}}
  \vspace{-2mm}
\end{table}

\textcolor{blue}{We supplement the comparison experiments in the noisy model homogeneous scenarios with numerous clients (Table~\ref{tab:numclient_homo_cifar10}\&\ref{tab:numclient_homo_cifar100}). Here, the local model architecture of all clients is set to ResNet12.
From the results we can see that our method is very competitive even in noisy model homogeneous federated learning scenarios. Especially when the local dataset is Cifar10, our method achieves higher accuracy than other SOTA methods. }

\end{appendix}

\end{sloppypar}